\documentclass{article}
\usepackage{ijcai22}

\usepackage{times}
\usepackage{soul}
\usepackage{url}
\usepackage[hidelinks]{hyperref}
\usepackage[utf8]{inputenc}
\usepackage[small]{caption}
\usepackage{graphicx}
\usepackage{amsmath}
\usepackage{amsthm}
\usepackage{booktabs}
\usepackage{algorithm}
\usepackage{algorithmic}
\usepackage{booktabs}
\usepackage{amsmath}
\usepackage{algorithm}
\usepackage{algorithmic}
\usepackage{tabularx}
\usepackage{amsfonts}
\usepackage{float}
\usepackage{subcaption}
\usepackage{longtable}
\usepackage{appendix}

\title{CCLF: A Contrastive-Curiosity-Driven Learning Framework for Sample-Efficient Reinforcement Learning}

\author{
	Chenyu Sun$^{1,2,3}$
	\and
	Hangwei Qian$^{2,4}$\footnote{Co-corresponding authors}\And
	Chunyan Miao$^{1,2,4*}$ 
	\affiliations
	$^1$Alibaba-NTU Singapore Joint Research Institute\\
	$^2$School of Computer Science and Engineering, Nanyang Technological University\quad$^3$Alibaba Group\\
	$^4$Joint NTU-UBC Research Centre of Excellence in Active Living for the Elderly (LILY)
	\emails
	\{chenyu002, qian0045\}@e.ntu.edu.sg,
	ascymiao@ntu.edu.sg
}

\begin{document}

\maketitle

\begin{abstract}
	
	
In reinforcement learning (RL), it is challenging to learn directly from high-dimensional observations, where data augmentation has recently been shown to remedy this via encoding invariances from raw pixels. Nevertheless, we empirically find that not all samples are equally important and hence simply injecting more augmented inputs may instead cause instability in Q-learning. In this paper\footnote{Accepted by IJCAI 2022}, we approach this problem systematically by developing a model-agnostic \textbf{C}ontrastive-\textbf{C}uriosity-driven \textbf{L}earning \textbf{F}ramework (CCLF), which can fully exploit sample importance and improve learning efficiency in a self-supervised manner. Facilitated by the proposed contrastive curiosity, CCLF is capable of prioritizing the experience replay, selecting the most informative augmented inputs, and more importantly regularizing the Q-function as well as the encoder to concentrate more on under-learned data. Moreover, it encourages the agent to explore with a curiosity-based reward. As a result, the agent can focus on more informative samples and learn representation invariances more efficiently, with significantly reduced augmented inputs. We apply CCLF to several base RL algorithms and evaluate on the DeepMind Control Suite, Atari, and MiniGrid benchmarks, where our approach demonstrates superior sample efficiency and learning performances compared with other state-of-the-art methods. 
\end{abstract}

\section{Introduction}
Despite the success of reinforcement learning (RL), extensive data collection and environment interactions are still required to train the agents \cite{laskin2020reinforcement}. In contrast, human beings are capable of learning new skills quickly and generalizing well with limited practice. Therefore, bridging the gap of sample efficiency and learning capabilities between machine and human learning has become a main challenge in the RL community \cite{rakelly2019efficient,schwarzer2020data,yarats2021improving,pmlr-v139-malik21c,sun2022psychological}. 

This challenge is particularly vital in learning directly from raw pixels. More recently, data augmentation methods are leveraged to incorporate more invariances, promote data diversity, and thereby enhance representation learning  \cite{laskin2020reinforcement,yarats2020image,yarats2021drqv2}.
Ideally, injecting a larger number of augmented samples should lead to a better model with invariances. Nevertheless, a noticeable trade-off is the computational complexity introduced. What's worse, simply increasing the number of augmented inputs may alter the semantics of samples, which has been empirically shown in our experimental results.
Moreover, the samples used for data augmentation are uniformly drawn from the replay buffer \cite{laskin2020curl,laskin2020reinforcement,yarats2020image,yarats2021drqv2}, which is inefficient as they are not equally important to learn.
These assumptions deviate from human-like intelligence, where 
humans can learn efficiently by curiously focusing on novel knowledge and revisiting old knowledge less frequently. Therefore, replaying the most under-explored experiences and selecting the most informative augmented inputs are the keys to improving sample efficiency and learning capability. 

To tackle these challenges, we propose a \textbf{C}ontrastive-\textbf{C}uriosity-driven \textbf{L}earning \textbf{F}ramework (CCLF) by introducing contrastive curiosity into four important components of RL including experience replay, training input selection, learning regularization, and task exploration without much computational overhead. 
Inspired by the psychological curiosity that can be externally stimulated, encompassing complexity, novelty, and surprise \cite{berlyne1960conflict,spielberger2012curiosity,liquin2020explanation}, we define the contrastive curiosity based on the surprise conceptualized by the agent's internal belief towards the augmented inputs. The internal belief is modeled by reusing the contrastive loss term in CURL \cite{laskin2020curl}, which can quantitatively measure the curiosity level without introducing any additional network architecture. 
With the proposed contrastive curiosity, agents can sample more under-explored transitions from the replay buffer, and select the most informative augmented inputs to encode invariances. This process can significantly reduce the amount of data used in RL without sacrificing the invariances. 
Thereafter, CCLF further utilizes the contrastive curiosity to regularize both Q-function and encoder by concentrating more on the surprising inputs, and intrinsically rewards agents for exploring under-learned observations. 

Our contribution can be highlighted as follows. 1)~We empirically demonstrate that not all samples nor their augmentations are equally important in RL. Thus, agents should learn curiously from the most important ones in a self-supervised manner. 
2)~A surprise-aroused type of curiosity, namely, contrastive curiosity, is proposed by reusing the representation learning module without increasing the network complexity.
3)~The proposed CCLF is capable of improving the sample efficiency and adapting the learning process directly from raw pixels, where the contrastive curiosity is fully exploited in different RL components in a self-navigated and coherent way. 
4) CCLF is model-agnostic and can be applied to model-free off-policy and on-policy RL algorithms. 
5)~Compared to other approaches, CCLF obtains state-of-the-art performance on the DeepMind Control (DMC) suite \cite{tunyasuvunakool2020dm_control}, Atari Games \cite{bellemare2013arcade}, and MiniGrid \cite{gym_minigrid} benchmarks.

\section{Related Works}
\paragraph{Data Augmentation in Sample-Efficient RL.}
Data augmentation has been widely applied in computer vision but is only recently introduced in RL to incorporate invariances for representation learning \cite{laskin2020reinforcement,yarats2020image,laskin2020curl}. 
To further improve the sample efficiency, one approach is to automatically apply the most effective augmentation method on any given task, through a multi-armed bandit or meta-learning the hyper-parameters to adapt \cite{raileanu2020automatic}. However, the underlying RL algorithm can become non-stationary and it costs more time to converge. Another approach is to regularize the learning process with observations from different training environments \cite{wang2020improving} or different steps \cite{yarats2021drqv2}. By injecting greater perturbations from other tasks and steps, the encoded features and learned policies can become more robust to task invariances. 
Different from these works, the proposed CCLF primarily focuses on perturbations generated in a single task and step, and selects the most under-learned transition tuples and their augmented inputs. 
As not all samples nor their augmented inputs are equally important, our work exploits the sample importance to adapt the learning process by concentrating more on under-explored samples. Most importantly, the amount of augmentations can be greatly reduced and the sample efficiency is improved without introducing complicated architectures. 

\paragraph{Curiosity-Driven RL.}
In curiosity-driven RL, agents are intrinsically motivated to explore the environment and perform complex control tasks by incorporating curiosity \cite{aubret2019survey,sun2022psychological}. In particular, curiosity is mainly used as a sophisticated intrinsic reward based on state novelty \cite{bellemare2016unifying}, state prediction errors \cite{pathak2017curiosity}, and uncertainty about outcomes \cite{li2021mural} or environment dynamics \cite{seo_chen2021re3}. Meanwhile, it can also be employed to prioritize the experience replay towards under-explored states \cite{schaul2016prioritized,zhao2019curiosity}. However, additional networks are required to model curiosity, which can be computationally inefficient and unstable for high-dimensional inputs with continuous controls. Moreover, none of these works have yet attempted to improve the sample efficiency and resolve the instability caused by data augmentation. CCFDM \cite{nguyen2021sample} is a concurrent work that incorporates CURL with action embedding and forward dynamics to formulate an intrinsic reward. Different from CCFDM, our framework does not require any additional architecture but only reuses the contrastive term in CURL that predicts more stably. More importantly, the proposed CCLF seamlessly integrates the curiosity mechanism into experience replay, training input selection, learning regularization, and environment exploration to concentrate more on under-learned samples and improve the sample efficiency stably.

\begin{figure*}
	\centering	
	\vspace{-4.5mm}
	\includegraphics[width=0.73\textwidth]{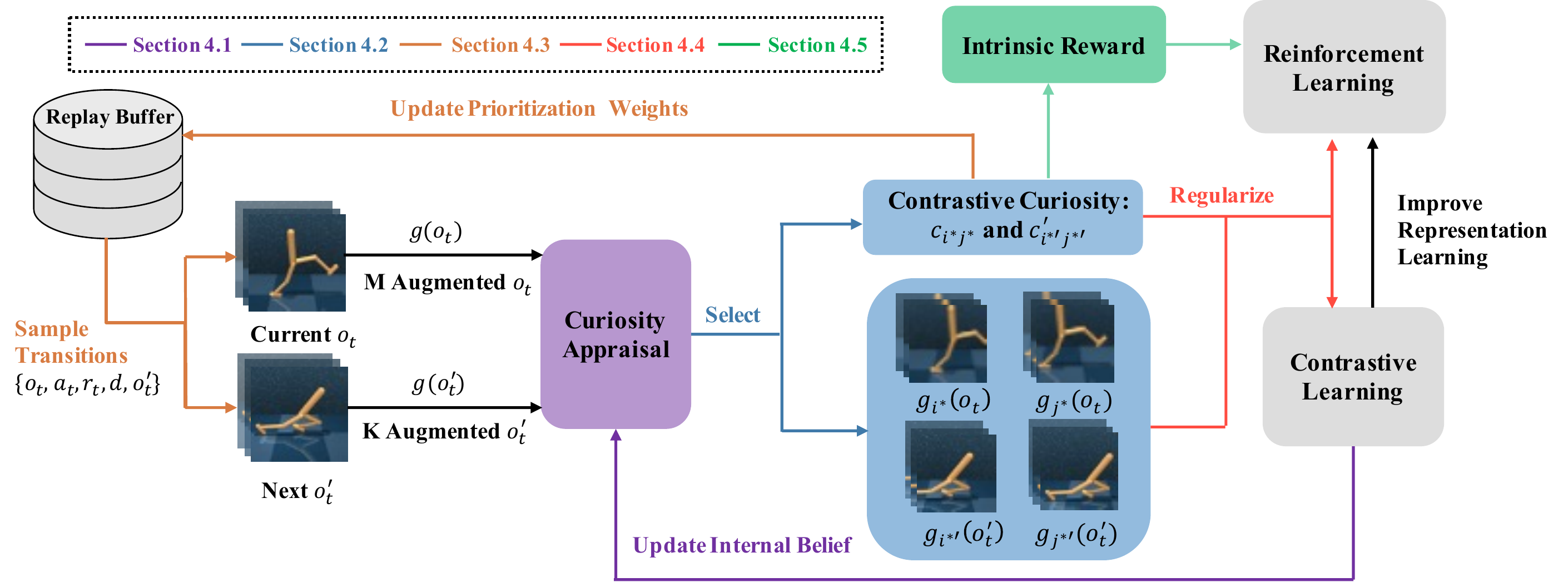} 
	\vspace{-1.8mm}
	\caption{\textbf{C}ontrastive-\textbf{C}uriosity-driven \textbf{L}earning \textbf{F}ramework (CCLF): a batch of transitions are sampled w.r.t. their prioritization weights. Image augmentation is performed to obtain $M$ augmented $o_t$ and $K$ augmented $o'_t$. The curiosity appraisal module quantitatively evaluates the contrastive curiosity and selects the two most informative inputs for both current and next observations. More importantly, the contrastive curiosity is simultaneously utilized to update the prioritization weights, construct an intrinsic reward, and adaptively regularize the contrastive learning and Q-learning modules. The contrastive learning module improves the representation learning and updates the agent's internal belief.}
	\label{workflow}
	\vspace{-2.8mm}
\end{figure*}

\section{Background}
In this paper, we consider a Markov Decision Process (MDP) setting with the state $s_t \in \mathcal{S}$, the action $a_t \in \mathcal{A}$, the transition probability $P$ mapping from the current state $s_t$ and action $a_t$ to the next state $s'_{t}$, and the (extrinsic) reward $r^e_t \in \mathcal{R}$. More details are provided in Appendix~\ref{app:extended_background}.
\paragraph{Soft Actor-Critic (SAC).}
\label{section:SAC}
SAC \cite{haarnoja2018soft} is an off-policy model-free algorithm that learns a stochastic policy $\pi_\psi$ (actor) with state-action value functions $Q_{\phi_1}, Q_{\phi_2}$ (critics), and a temperature $\alpha$ by encouraging exploration through a $\gamma$-discounted maximum-policy-entropy term. However, agents are often required to learn directly from high-dimensional observations $o_t \in \mathcal{O}$ rather than states $s_t$ in practice. In this paper, we demonstrate our framework mainly using SAC with raw pixel inputs as the base algorithm.

\paragraph{Contrastive Unsupervised RL (CURL).}
\label{Contrastive Learning}
CURL \cite{laskin2020curl} utilizes data augmentation and contrastive learning to train an image encoder $f_\theta(o)$ in a self-supervised way, imposing an instance discrimination between similar ($^+$) and dissimilar ($^-$) encoded states. Given a batch of visual observations $o$, each is augmented twice and encoded into a query $q=f_\theta(o_q)$ and a key $k=f_{\bar{\theta}} (o_k)$. The key encoder $f_{\bar{\theta}}$ is a momentum-based moving average of the query encoder $f_\theta$ to ensure consistency and stability, and $f_\theta$ is learned by enforcing $q$ to match with $k^+$ while keeping far apart from $k^-$. 

\paragraph{Data Regularized Q-Learning (DrQ).}
Based on SAC settings, DrQ \cite{yarats2020image} incorporates optimality invariant image transformations to regularize the Q-function, improving robust learning directly from raw pixels. Let $g(o)$ represent the random image crop augmentation on observations $o$. It should ideally preserve the Q-values s.t. $Q(o, a) = Q(g_i(o), a), \forall o \in \mathcal{O}, a \in \mathcal{A}, i = 1,2,3,\cdots$.
DrQ then applies data augmentation to each transition tuple $\tau_t=(o_t, a_t, r^e_t, d_t, o'_{t})$ that is uniformly sampled from the replay buffer $\mathcal{B}$, where $d_t$ is the done signal. With $K$ augmented next observations $g_k(o'_{t})$ and $M$ augmented current observations $g_m(o_t)$, the critic $Q_\phi$ can be regularized by averaging over $M$ augmented inputs from $o_t$,
\begin{equation}
\label{Q2}
\resizebox{0.9\hsize}{!}{$
	\mathcal{L}_Q(\phi)= \mathbb{E}_{\tau \sim \mathcal{B}} \left[ \frac{1}{M}\sum_{m=1}^{M} \Big( Q_\phi (g_{m}(o_t), a_t) \\- (r^e_t+\gamma (1-d_t)\mathcal{T}_t)\Big)^2 \right]$}
\end{equation}
where $\mathcal{T}_t$ is the soft target value and it can also be regularized by averaging over $K$ augmented inputs from $o'_t$, 
\begin{equation}
\label{Q1}
\resizebox{0.9\hsize}{!}{$
\mathcal{T}_t=\frac{1}{K}\sum_{k=1}^{K} \left[\min_{i=1,2}Q_{\bar{\phi_i}}(g_{k}(o'_{t}),a'_k)-\alpha \log \pi_\psi (a'_k | g_{k}(o'_{t}))\right]$.}
\end{equation} 

\section{The Proposed CCLF}
\textbf{C}ontrastive-\textbf{C}uriosity-driven \textbf{L}earning \textbf{F}ramework (CCLF) extends the model-free RL to further improve sample efficiency when learning directly from the raw pixels. In particular, it fully exploits sample importance for agents to efficiently learn from the most informative data. Firstly, we re-purpose the contrastive term in CURL \cite{laskin2020curl} without additional architectures to quantify the contrastive curiosity (Section~\ref{section:curiosity}). Subsequently, this contrastive curiosity is coherently integrated into four components to navigate RL with minimum modification: augmented input selection (Section~\ref{section:select}), experience replay (Section \ref{section:priority}), Q-function and encoder regularization (Section~\ref{section:regularize}), and environment exploration (Section~\ref{section:reward}), as illustrated in Figure~\ref{workflow}.
Without loss of generosity, we apply CCLF on the state-of-the-art off-policy RL algorithm, SAC \cite{haarnoja2018soft} as summarized in Algorithm~\ref{alg:algorithm}\footnote{Our code is available at \url{https://github.com/csun001/CCLF}.}. Extensions on other base algorithms are carried out in Section \ref{additonal_experiments} and Appendix~\ref{app:rainbow} and~\ref{app:A2C}.


\subsection{Contrastive Curiosity}
\label{section:curiosity}
Curiosity can be aroused by an unexpected stimulus that behaves differently from the agent's internal belief. 
To quantify such surprise-aroused curiosity, we define the agent's curiosity $c_{ij}$ by the prediction error of whether any two augmented observations $g_{i}(o), g_{j}(o)$ are from the same observation $o$,
\begin{equation}
\label{curiosity}
c_{ij} = 1 - \text{IB}(g_{i}(o), g_{j}(o)) \in [0,1]
\end{equation}
where IB represents agent's internal belief of whether $g_{i}(o), g_{j}(o)$ are augmented (\textit{e.g.}, randomly cropped) from the same $o$ with similar representations. Since the contrastive loss can be viewed as the log-loss of a softmax-based classifier to match a query $q$ with the key $k$ from the same observation in a batch, it becomes a natural choice for measuring agent's internal belief \small 
$
\text{IB}(g_{i}(o), g_{j}(o))=\frac{\exp(q^TWk^+)}{\exp(q^TWk^+)+\sum_{l=1}^{B-1}\exp (q^TWk_{l}^-)}
$, \normalsize where $B$ is the batch size and $q$ is the query encoder \small $q=f_\theta(g_{i}(o))$. \normalsize
Moreover, we denote the key encoder \small $k=f_{\bar{\theta}}(g_{j}(o))$ \normalsize as $k^+$ if its input is the same as that in the query encoder $q$; otherwise, we denote it as $k^-$.
An immediate merit is that the contrastive curiosity does not require any additional architecture or auxiliary loss because IB is updated directly through representation learning in a self-supervised way.

A higher contrastive curiosity value indicates that the agent does not believe $q$ is similar to $k^+$ or the agent mistakenly matches $q$ with some $k^-$ instead, which ultimately results in a surprise in a self-supervised way. It further implies that the sampled transition tuple contains novel information that has yet been learned by the agent, and the encoder $f_\theta$ is not optimal to extract a meaningful state representation from raw pixels. 
With the proposed contrastive curiosity in-place, we can integrate different curiosity-driven mechanisms in the proposed CCLF to achieve sample-efficient RL, which is discussed in the following sub-sections.

\subsection{Curiosity-Based Augmented Input Selection}
\label{section:select}
Although DrQ~\cite{yarats2020image} has shown that increasing~$[K,M]$, \textit{i.e.}, the amounts of augmented inputs on next and current observations respectively, can potentially improve agent’s performances through regularized Q-learning, a crucial trade-off is the introduced higher computational complexity. In addition, more augmented data does not necessarily lead to better performance, as data transformations might alter the semantics and result in the counterproductive performance. To tackle these challenges, we aim to select the most informative inputs for the subsequent learning. Without loss of generality, we assume two inputs are selected from $M$ augmented current observations $o_t$ and similarly two are selected from $K$ augmented next observations $o'_{t}$ in this paper.

It should be noted that there are various ways to select the most informative augmented inputs, where one straightforward way is to select by least overlap in pixels,
\begin{equation}
\label{select_pixel}
i^*,j^* =\text{arg}\min \text{Overlap} (g_{i}(o), g_{j}(o)) \ \forall i, j, i \neq j.
\end{equation}
However, a more human-like way is to select the most representative inputs based on the curious level conceptualized by the internal belief rather than simple visual overlaps. Therefore, we propose to select the augmented inputs that cause highest contrastive curiosity as defined in Eq. (\ref{curiosity}), 
\begin{equation}
\label{max_curiosity}
i^*,j^* =\text{arg}\max c_{ij} \ \forall i, j, i \neq j.
\end{equation}
In this way, the augmented inputs that are most challenging for matching can be curiously identified since they potentially contain novel knowledge that has yet been learned; meanwhile, this selection mechanism can help to encode more representative state information from the selected inputs, while the agent's internal belief can be jointly updated. 
As a result, an improved encoder that is robust to different views of observations can be trained with fewer inputs, potentially yielding an improvement for the sample efficiency.


\subsection{Curiosity-Based Experience Replaying}
\label{section:priority}
In the conventional off-policy RL, agents uniformly sample transitions $\tau$ from the replay buffer to learn policies. 
Although they can eventually perform a complex task by repeatedly practicing in a trial-and-error fashion, we hypothesize that a more sample-efficient and generalizable way is to revisit the transitions that are relatively new or different more frequently. 
Therefore, we prioritize the experience replay by assigning different prioritization weights $w \in [0,1]$ to all transitions stored in the replay buffer $\mathcal{B}$. In particular, the prioritization weight is initialized to $w_0=1$ for any newly added transition tuple. Thereafter, we propose to update the weights of transitions with the overall contrastive curiosity at each training step $s$,
\begin{equation}
\label{sample_weight}
w_s = \beta w_{s-1} + \frac{1}{2}(1-\beta ) (c_{i^*j^*}+c_{i^{*\prime}j^{*\prime}}')
\end{equation}
where $\beta \in [0,1]$ is a momentum coefficient, and $c_{i^*j^*}, c_{i^{*\prime}j^{*\prime}}'$ are the contrastive curiosity about $o$ and $o'$ respectively. The intuition of the momentum update is to maintain a stable update such that the transitions arousing low curiosity will be gradually de-prioritized for learning. Mathematically, the probability of $\tau_i$ to be replayed is $p(\tau_i)=\frac{w_i}{\sum_{n=1}^{N} w_n}$ and it becomes small only when $\tau_i$ has been sampled many times. Hence, more recent and surprising transitions arousing high curiosity can be sampled more frequently to learn.

\subsection{Curiosity-Based Regularization}
\label{section:regularize}
Although agents can benefit from learning the selected complex inputs, it imposes challenges for agents as well, which may cause unstable and poor performances. Hence,
it is crucial to adapt the learning process by concentrating more on under-learned knowledge. To achieve this, we propose an adaptive regularization for both Q-function and the observation encoder, guided by the contrastive curiosity in order to learn more from the selected inputs arousing high curiosity. In particular, we modify Eq.~(\ref{Q1}) and Eq.~(\ref{Q2}) as
\begin{equation}
\footnotesize
\label{new_Q2}
\begin{aligned}
& \mathcal{L}_Q(\phi) = \mathbb{E}_{\tau \sim \mathcal{B}} \left[ (1-c_{i^*j^*}) \mathcal{E}_{i^*}^2+ c_{i^*j^*}  \mathcal{E}_{j^*}^2 \right],
\\[-0mm] &\mathcal{T}_t = (1-c_{i^{*\prime}j^{*\prime}}') \mathcal{T}^{i^{*\prime}}_t +c_{i^{*\prime}j^{*\prime}}' \mathcal{T}^{j^{*\prime}}_t,
\\[-0mm] &\text{where } \mathcal{E}_{m} =Q_\phi (g_{m}(o_t), a_t) - (r^e_t+\gamma (1-d_t)\mathcal{T}_t), \; m=i^{*}, j^{*},
\\[-0mm] &\text{and } \mathcal{T}^{k}_t =\min_{l=1,2} Q_{\bar{\phi_l}}(g_{k}(o'_{t}),a'_k)-\alpha \log \pi_\psi (a'_k | g_{k}(o'_{t})), k=i^{*\prime},j^{*\prime}.
\end{aligned}
\end{equation}
\begin{table*}[h!]
	 \vspace{-1.5mm}
	\small
	\begin{tabular}{@{}p{3.2cm}p{0.08\textwidth}p{0.08\textwidth}p{0.08\textwidth}p{0.08\textwidth}p{0.08\textwidth}p{0.08\textwidth}p{0.08\textwidth}|p{0.08\textwidth}@{}}
		\toprule
		\small
		100K Step   Scores   & SAC-Pixel   & CURL                & DrQ                 & CURL+       & CURL++      & Select      & Select+            & 
		CCLF                \\ \midrule
		Finger, Spin       & 230$\pm$194 & 686$\pm$113 & 784$\pm$173 & 780$\pm$96  & 735$\pm$120 & 699$\pm$138 & 768$\pm$90  & 
	    \textbf{944$\pm$42}  \\
		Cartpole, Swingup  & 237$\pm$49  & 524$\pm$179 & 675$\pm$174 & 694$\pm$87  & 665$\pm$122 & 624$\pm$182 & 561$\pm$181 & \textbf{799$\pm$61} \\
		Reacher, Easy      & 239$\pm$183 & 566$\pm$226 & 682$\pm$86  & 541$\pm$190 & 479$\pm$216 & 646$\pm$171 & 616$\pm$284 & 
		\textbf{738$\pm$99} \\
		Cheetah, Run       & 118$\pm$13  & 286$\pm$65  & \textbf{332$\pm$36}  & 302$\pm$50  & 264$\pm$53  & 251$\pm$26  & 265$\pm$69  & 317$\pm$38  \\
		Walker Walk        & 95$\pm$19   & 482$\pm$237 & 492$\pm$267 & 484$\pm$61  & 504$\pm$142 & 453$\pm$91  & 408$\pm$170 & \textbf{648$\pm$110} \\
		Ball in Cup, Catch & 85$\pm$130  & 667$\pm$197 & 828$\pm$131 & 687$\pm$260 & 728$\pm$143 & 732$\pm$223 & 739$\pm$132 & \textbf{914$\pm$20} \\ \midrule
		500K Step   Scores   & SAC-Pixel   & CURL                & DrQ                 & CURL+       & CURL++      & Select      & Select+            & CCLF                 \\ \midrule
		Finger, Spin       & 346$\pm$95  & 783$\pm$192 & 803$\pm$198 & 855$\pm$164 & 838$\pm$164 & 803$\pm$167 & 879$\pm$153 & \textbf{974$\pm$6}   \\
		Cartpole, Swingup  & 330$\pm$73  & 847$\pm$28  & 858$\pm$19  & 853$\pm$22  & 852$\pm$17  & 855$\pm$26  & 837$\pm$38  & 
		\textbf{869$\pm$9}   \\
		Reacher, Easy      & 307$\pm$65  & 
		\textbf{956$\pm$40}  & 939$\pm$44  & 933$\pm$62  & 937$\pm$40  & 939$\pm$78  & 906$\pm$80  & 941$\pm$48  \\
		Cheetah, Run       & 85$\pm$51   & 440$\pm$144 & 536$\pm$115 & 518$\pm$24  & 495$\pm$97  & 417$\pm$59  & 470$\pm$78  & 
	    \textbf{588$\pm$22}  \\
		Walker Walk        & 71$\pm$52   & 928$\pm$26  & 887$\pm$126 & 916$\pm$27  & 914$\pm$24  & 921$\pm$27  & 850$\pm$64  & \textbf{936$\pm$23}  \\
		Ball in Cup, Catch & 162$\pm$122 & \textbf{956$\pm$14}  & \textbf{956$\pm$14}  & 951$\pm$19  & \textbf{956$\pm$8}   & 949$\pm$21  & 949$\pm$24  & \textbf{961$\pm$9}  \\ \bottomrule
	\end{tabular}
    \vspace{-2mm}
	\caption{Performance scores (mean \& standard deviation) on DMC evaluated at 100K and 500K environment steps. CCLF outperforms other approaches on 5 out of 6 tasks in both sample efficiency (100K) and asymptotic performance (500K) regimes, across 6 random seeds. 
	}
   \vspace{-2.5mm}
	\label{table:results}
\end{table*}
\normalsize
It is worth noting that this regularized Q-function is rather general to recover other state-of-the-art works as special cases.
When all augmented inputs arouse exactly moderate level of curiosity $c_{i^*j^*}=c_{i^{*\prime}j^{*\prime}}'=\frac{1}{2}$, the proposed regularization is equivalent to DrQ with $[K,M]=[2, 2]$. Moreover, when the agent can perfectly match the two augmented inputs with no contrastive curiosity $c_{i^*j^*}=c_{i^{*\prime}j^{*\prime}}'=0$,  it is sufficient to update the Q-function with only one input; when the agent fails to encode any similarity and becomes extremely curious $c_{i^*j^*}=c_{i^{*\prime}j^{*\prime}}'=1 $, it should focus completely on the novel input instead. Both cases can reproduce the work of RAD \cite{laskin2020reinforcement}. 
Most importantly, our proposed Q-function regularization enables the agent to adapt the learning process in a self-supervised way that it is fully controlled by the conceptualized contrastive curiosity to exploit sample importance and stabilize the learning process. 

Similarly, we also regularize the representation learning in a curious manner, inspired by the solution to the supervised class imbalance problem. To deal the with training set containing under-represented classes, a practical approach is to inversely weight the loss of each class according to their sizes. We follow this motivation to incorporate the contrastive curiosity $c^b_{i^*j^*}$ about the current observations as the weight for each log-loss class $b$ to update the encoder $f_{\theta}$,
\begin{equation}
\label{proposed_contrastive_loss}
\resizebox{0.89\hsize}{!}{$
\mathcal{L}_f(\theta) = -\sum_{b=1}^{B} c^b_{i^*j^*} \log \frac{\exp(q_b^TWk^+)}{\exp(q_b^TWk^+)+\sum_{l=1}^{B-1}\exp (q_b^TWk_l^-)} $}
\end{equation}
where samples arousing high contrastive curiosity will be considered as under-represented classes and therefore agents need to adaptively pay more attention during the representation learning by optimizing $f_\theta$. Meanwhile, agents also jointly re-calibrate a proper internal belief by updating $W$.
\begin{algorithm}[tb]
	\caption{An Implementation of CCLF on SAC}
	\label{alg:algorithm}
	\textbf{Input}: MDP $\tau_t = (o_t, a_t, r^e_t, d_t, o'_{t})$, numbers of augmented inputs $[K,M]$, replay buffer $\mathcal{B}$, training step $T$, batch size $B$ \\
	\textbf{Parameter}: Observation encoder network~$\theta$, actor network~$\psi$, critics networks~$\phi_i$, temperature coefficient~$\alpha$, and bilinear product weight~$W$ \\
	\textbf{Output}: Optimal policy $\pi_\psi^*$
	\begin{algorithmic}[0] 
		\FOR{$t=1 $ \textbf{to} $T$}
		\STATE $a_t \sim \pi_\psi(\cdot|g(o_t))$
		\STATE $\mathcal{B} \cup (o_t, a_t, r^e_t, d_t, o'_{t}) \to \mathcal{B}$ with $w_t=1$
		\STATE Sample a minibatch $\{(o_l, a_l, r^e_l, d_l, o'_{l})\}_{l=1}^{B} \stackrel{w_l}{\sim} \mathcal{B}$ based on the prioritization weight $w_l$
		\FOR{each sample $\tau_l$ in the minibatch}
		\STATE Augment $o_l$ and $o'_{l}$ via $g(\cdot)$ to obtain $M$ and $K$ inputs\\ 
		\STATE Evaluate the contrastive curiosity $c_{ij}$ by Eq. (\ref{curiosity})  
		\STATE Select $g_{i^*}(o_l), g_{j^*}(o_l)$ from $M$ augmented $o_l$ and select  $g_{i^{*\prime}}(o'_{l}), g_{j^{*\prime}}(o'_{l})$ from $K$ augmented $o'_l$ by Eq.~(\ref{max_curiosity})\\
		\STATE $r_l=r^e_l + r^i_l$ with $r_l^i$ from Eq. (\ref{intrinsic})
		\STATE Update $w_l$ according to Eq. (\ref{sample_weight})
		\ENDFOR
		\STATE Update critics $Q_{\phi_i}$ by Eq.~(\ref{new_Q2})
		\STATE Update the actor $\pi_\psi$ and temperature coefficient $\alpha$
		\STATE Update encoder $f_\theta$ and $W$ by Eq. (\ref{proposed_contrastive_loss})
		\STATE $o_{t+1}=o'_{t}$
		\ENDFOR
	\end{algorithmic}
\end{algorithm}

\subsection{Curiosity-Based Exploration}
\label{section:reward}
Intrinsic rewards can motivate agents to explore actively \cite{sun2022psychological}, improving the sample efficiency in the conventional RL. While SAC alone can be viewed as the entropy maximization of agent's actions intrinsically, in the proposed CCLF, we explicitly define an intrinsic reward proportional to the average contrastive curiosity about $o_t$ and $o'_{t}$,
\begin{equation}
\label{intrinsic}
r_t^i= \lambda \exp (-\eta t) \frac{r^e_{max}}{r^i_{max}}\frac{c_{i^*j^*}+c_{i^{*\prime}j^{*\prime}}'}{2}
\end{equation}
where $\lambda$ is a temperature coefficient, $\eta$ is a decay weight, $t$ is the environment step, $r^e_{max}$ and $r^i_{max}$ are respectively the maximum extrinsic and intrinsic rewards over step $t$.

With the proposed $r_t^i$ to supplement $r_t^e$ in Eq.~(\ref{new_Q2}), agents can be encouraged to explore the surprising states that arouse high contrastive curiosity substantially. In particular, higher $r_t^i$ rewards agents for exploration when different views of the same observations produce inconsistent representations. Meanwhile, $r_t^i$ is decayed with respect to the environment step $t$ to ensure the convergence of policies. As the extrinsic reward $r^e$ differs across different tasks, the normalization is performed to balance $r^e$ and $r^i$. This formulation is similar to the intrinsic reward in CCFDM \cite{nguyen2021sample}, but the proposed CCLF does not require a forward dynamic model or action embedding that increases the model complexity. 



\begin{figure*}[t]
	\centering
		\vspace{-2mm}
	\includegraphics[width=1\textwidth]{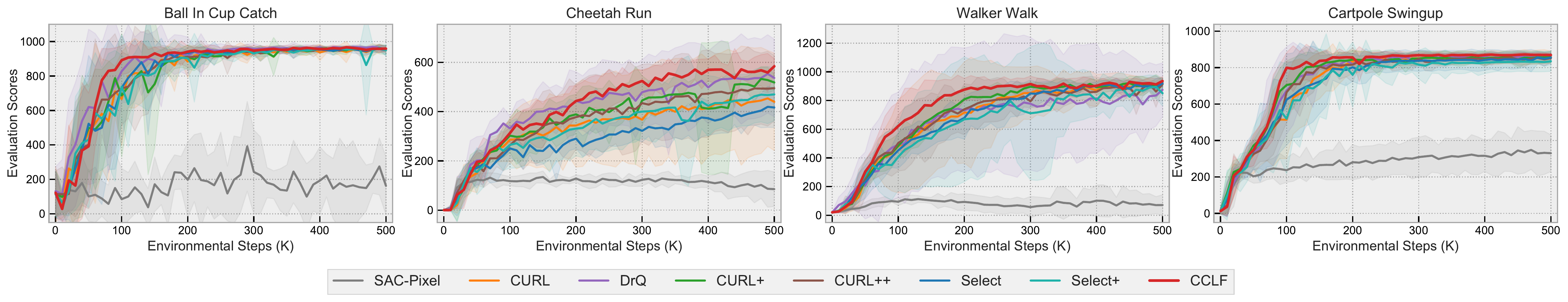} 
	\vspace{-6mm}
	\caption{Learning performances on the continuous control tasks from the DMC Suite (Selected). The proposed CCLF on SAC outperforms the other baseline methods in terms of sample efficiency and converges much faster, averaged by 6 random runs. 
	}
	\label{fig:results}
	\vspace{-3mm}
\end{figure*}
\begin{figure*}[t]
	\centering
	\vspace{-1mm}
	\includegraphics[width=1\textwidth]{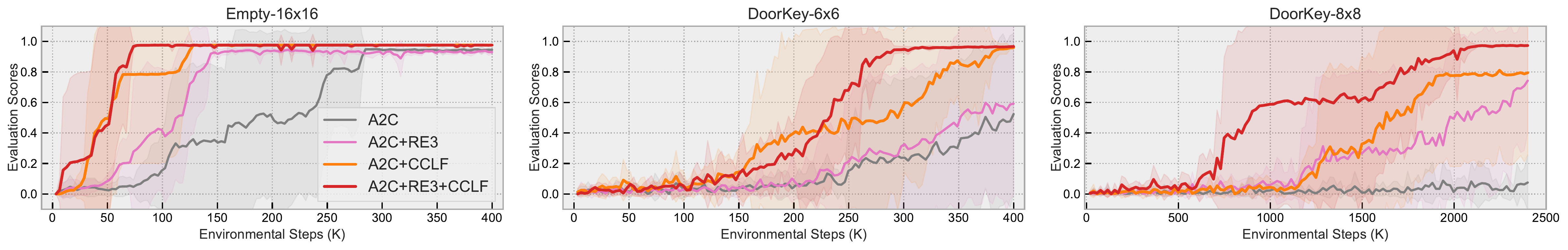} 
	\vspace{-6mm}
	\caption{Learning performances on the navigation tasks from the MiniGrid. The proposed CCLF can be applied to both A2C and A2C+RE3. It significantly outperforms the other baselines in terms of sample efficiency and converges much faster, averaged by 5 random runs.  
	}
	\label{fig:A2Cresults}
	\vspace{-4mm}
\end{figure*}

\section{Experiments and Results}
\subsection{Experimental Setup}

\normalsize
We empirically evaluate the proposed CCLF in terms of sample efficiency and ultimate performance, on 6 continuous control tasks from the DMC suite~\cite{tunyasuvunakool2020dm_control}, 26 discrete control tasks from the Atari Games \cite{bellemare2013arcade} and 3 navigation tasks with sparse extrinsic rewards from the MiniGrid \cite{gym_minigrid}. In this section, we mainly present the experimental results in the DMC suite with SAC being the base algorithm while detailed settings and results in the Atari Games and the MiniGrid are included in Appendix~~\ref{app:rainbow}, ~\ref{app:A2C},~\ref{app:rainbow_results},~and~\ref{app:A2C_results}. 
For a comprehensive evaluation in the DMC suite, we include the following baselines to compare against: 
\begin{itemize}
	\item Pixel-based SAC (SAC-Pixel) \cite{haarnoja2018soft}
	\item CURL \cite{laskin2020curl}. 
	\item DrQ \cite{yarats2020image} with $[K,M]=[2, 2]$ and a modified augmentation method for consistency.
	\item Hybrids of CURL and DrQ: CURL+ and CURL++, where contrastive representation learning is integrated to DrQ for $[K,M]=[2, 2]$ and $[5, 5]$ respectively.
	\item Augmented input selection models: 2 out of 5 inputs for each sample are selected by pixel overlap (Select) via Eq. (\ref{select_pixel}) and contrastive curiosity (Select+) via Eq. (\ref{max_curiosity}) without the other curiosity-based components.
\end{itemize}
The detailed setting of hyper-parameters is provided in Appendix~\ref{app:SAC_settings}. For our proposed CCLF, we initialize it with $[K,M]=[5, 5]$ to generate a sufficiently large amount of augmented inputs. For simplicity, we fix $i^*$ randomly and only select $j^*$ via Eq.~(\ref{max_curiosity}) for the augmented input selection. 


\subsection{Results and Discussion}
\paragraph{Not all Samples are Equally Important.}
In CURL+, data augmentation is applied twice for each sampled transition while 5 times in CURL++. Since CURL++ injects 2.5$\times$ larger amount of inputs than CURL+, its computational complexity increases dramatically. Table \ref{table:results} shows that CURL++ performs worse than CURL+ in 4 tasks at 100K steps and slightly outperforms CURL+ in only 2 tasks at 500K steps. In Figure~\ref{fig:results} and Appendix Figure~\ref{fig:SACresults}, the learning curve of CURL++ is clearly below CURL+ at first and gradually approaches to the same level as CURL+. Since more augmented inputs may not guarantee the consistency of semantics, additional training is often required for convergence. Therefore, we can empirically validate the hypothesis that not all augmented inputs are equally important and simply increasing the number of augmentations is instead inefficient. A similar result can be found DrQ Appendix F~\cite{yarats2020image}.

\subsubsection{Main Results on the DMC Suite}
The average sample efficiency and asymptotic performance are shown in Table~\ref{table:results} at 100K and 500K environment steps, respectively. Meanwhile, Figure~\ref{fig:results} demonstrates the agent's learning capability over 500K steps. Compared SAC-Pixel to the other models in Figure~\ref{fig:results}, its performance is not improved in all 6 tasks even until 500K steps while the other models can asymptotically perform well. Thus, it is challenging for the conventional SAC to learn directly from raw pixels and a sample-efficient RL method is needed to aid that.

According to Table~\ref{table:results}, Select performs better than Select+, on 3 tasks at 100K and 4 tasks at 500K, with more stable learning curves as shown in Figure~\ref{fig:results}. Indeed, the inputs in Select may contain some invariances to improve sample efficiency and learning capability. However, more under-learned inputs with even richer invariances are present in Select+, and agents cannot adapt the learning process in this model, causing the instability issue in Select+. 

To tackle this issue, the proposed CCLF collaboratively adapts the learning process with the selected inputs and contrastive curiosity, so the learning curves become more smooth than others in Figure~\ref{fig:results}. In particular, CCLF outperforms all baselines in 5 tasks at both 100K and 500K regimes in Table~\ref{table:results}. Moreover, it converges much faster than Select+ according to the results at 100K steps. In fact, the proposed CCLF only requires about 50\% environment steps to converge to desirable performances on 3 tasks (Ball in Cup, Walker, and Cartpole) as the other baselines. In addition, it even benchmarks on Cheetah-Run and Finger-Spin tasks at 500K steps. Therefore, we can conclude that our proposed CCLF can improve the sample efficiency and learning capabilities of RL agents, with fewer environment interactions and 60\% reduced augmented inputs. We also analyze the computational complexity on the Cartpole task by model sizes and training time in Appendix~\ref{app:computational_cost}, where CCLF can avoid increasing the training cost dramatically.

\paragraph{Additional Experiments in Atari Games.}
\label{additonal_experiments}
In addition to continuous control tasks, CCLF can also be incorporated into Rainbow DQN \cite{hessel2018rainbow} to perform discrete control tasks. As shown in Appendix~\ref{app:rainbow_results}, the proposed CCLF attains state-of-the-art performances in 8 out of 26 Atari Games at 100K steps. In particular, CCLF is superior to CURL in 11 games and DrQ in 18 games, which favorably indicates the effectiveness of improving sample efficiency. 

\paragraph{Further Investigation on MiniGrid.}
Apart from off-policy algorithms, we also investigate the compatibility on the on-policy algorithm. More specifically, we apply the proposed CCLF to A2C \cite{mnih2016asynchronous} and RE3 \cite{seo_chen2021re3} to perform navigation tasks with spares rewards in MiniGrid. We first adapt CCLF to the on-policy algorithm by removing the experience replay component. Note that the input from MiniGrid is already a compact and efficient $7\times7\times3$ embedding of partially-observable $7\times7$ grids, so even slight augmentation will induce highly inconsistent learned features. Thus, we directly duplicate the embedding without random augmentations to obtain contrastive curiosity for regularization and intrinsic reward. Figure~\ref{fig:A2Cresults} shows that CCLF exhibits superior sample efficiency and learning capabilities in all three tasks, even model-agnostic with the state-of-the-art curiosity-driven method RE3. In the Empty-16$\times$16 task, our CCLF can reach the optimal level in about 50\% and 55\% of the training steps of RE3 and A2C, respectively. By comparing the ultimate performance scores, the proposed CCLF obtains 1.63$\times$ higher average performance than RE3 in DoorKey-6$\times$6 and 1.3$\times$ in DoorKey-8$\times$8.


\paragraph{Effectiveness of the Proposed RL Components.}
One might wonder if the proposed CCLF benefits mainly from one or several curiosity-based components in practice. Hence, we empirically examine the effectiveness of all possible combinations of the four curiosity-driven components on the Cartpole task from the DMC suite. The results are included in Appendix~\ref{app:ablation}, where it can be concluded that all four components are necessary and important to attain state-of-the-art performances. Our proposed CCLF can navigate all four RL components together to improve the sample efficiency and resolve instability, which demonstrates effective collaboration.

\section{Conclusion }
In this paper, we present CCLF, a contrastive-curiosity-driven learning framework for RL with visual observations, which can significantly improve the sample efficiency and learning capabilities of agents. As we empirically find that not all samples nor their augmented inputs are equally important for RL, CCLF encourages agents to learn in a curious way, exploiting sample complexity and importance systemically. 



\section*{Acknowledgments}
This research is supported in part by the National Research Foundation, Prime Minister’s Office, Singapore under its NRF Investigatorship Programme (NRFI Award No. NRF-NRFI05-2019-0002). Any opinions, findings and conclusions or recommendations expressed in this material are those of the author(s) and do not reflect the views of National Research Foundation, Singapore. 
This research is supported in part by the Alibaba-NTU Singapore Joint Research Institute, Nanyang Technological University, and
in part by the Singapore Ministry of Health under its National Innovation Challenge on Active and Confident Ageing (NIC Project No. MOH/NIC/COG04/2017) and (NIC Project No. MOH/NIC/HAIG03/2017).
H.Qian thanks the support from the Wallenberg-NTU Presidential Postdoctoral Fellowship.
\small
\bibliographystyle{named}
\bibliography{ijcai22-appendix}

\begin{thebibliography}{}

\bibitem[\protect\citeauthoryear{Aubret \bgroup \em et al.\egroup
  }{2019}]{aubret2019survey}
Arthur Aubret, Laetitia Matignon, and Salima Hassas.
\newblock A survey on intrinsic motivation in reinforcement learning.
\newblock {\em arXiv:1908.06976}, 2019.

\bibitem[\protect\citeauthoryear{Bellemare \bgroup \em et al.\egroup
  }{2013}]{bellemare2013arcade}
Marc~G Bellemare, Yavar Naddaf, Joel Veness, and Michael Bowling.
\newblock The arcade learning environment: An evaluation platform for general
  agents.
\newblock {\em JAIR}, 47, 2013.

\bibitem[\protect\citeauthoryear{Bellemare \bgroup \em et al.\egroup
  }{2016}]{bellemare2016unifying}
Marc Bellemare, Sriram Srinivasan, Georg Ostrovski, Tom Schaul, David Saxton,
  and Remi Munos.
\newblock Unifying count-based exploration and intrinsic motivation.
\newblock {\em NeurIPS}, 29:1471--1479, 2016.

\bibitem[\protect\citeauthoryear{Berlyne}{1960}]{berlyne1960conflict}
Daniel~E Berlyne.
\newblock Conflict, arousal, and curiosity.
\newblock 1960.

\bibitem[\protect\citeauthoryear{Burda \bgroup \em et al.\egroup
  }{2018}]{burda2018exploration}
Yuri Burda, Harrison Edwards, Amos Storkey, and Oleg Klimov.
\newblock Exploration by random network distillation.
\newblock {\em arXiv preprint arXiv:1810.12894}, 2018.

\bibitem[\protect\citeauthoryear{Chevalier-Boisvert \bgroup \em et al.\egroup
  }{2018}]{gym_minigrid}
Maxime Chevalier-Boisvert, Lucas Willems, and Suman Pal.
\newblock Minimalistic gridworld environment for openai gym.
\newblock \url{https://github.com/maximecb/gym-minigrid}, 2018.

\bibitem[\protect\citeauthoryear{Cobbe \bgroup \em et al.\egroup
  }{2019}]{cobbe2019quantifying}
Karl Cobbe, Oleg Klimov, Chris Hesse, Taehoon Kim, and John Schulman.
\newblock Quantifying generalization in reinforcement learning.
\newblock In {\em ICML}, pages 1282--1289. PMLR, 2019.

\bibitem[\protect\citeauthoryear{Fortunato \bgroup \em et al.\egroup
  }{2017}]{fortunato2017noisy}
Meire Fortunato, Mohammad~Gheshlaghi Azar, Bilal Piot, Jacob Menick, Ian
  Osband, Alex Graves, Vlad Mnih, Remi Munos, Demis Hassabis, Olivier Pietquin,
  et~al.
\newblock Noisy networks for exploration.
\newblock {\em arXiv preprint arXiv:1706.10295}, 2017.

\bibitem[\protect\citeauthoryear{Haarnoja \bgroup \em et al.\egroup
  }{2018}]{haarnoja2018soft}
Tuomas Haarnoja, Aurick Zhou, Kristian Hartikainen, George Tucker, Sehoon Ha,
  Jie Tan, Vikash Kumar, Henry Zhu, Abhishek Gupta, Pieter Abbeel, et~al.
\newblock Soft actor-critic algorithms and applications.
\newblock {\em arXiv preprint arXiv:1812.05905}, 2018.

\bibitem[\protect\citeauthoryear{Hafner \bgroup \em et al.\egroup
  }{2019a}]{hafner2019dream}
Danijar Hafner, Timothy Lillicrap, Jimmy Ba, and Mohammad Norouzi.
\newblock Dream to control: Learning behaviors by latent imagination.
\newblock In {\em International Conference on Learning Representations}, 2019.

\bibitem[\protect\citeauthoryear{Hafner \bgroup \em et al.\egroup
  }{2019b}]{hafner2019learning}
Danijar Hafner, Timothy Lillicrap, Ian Fischer, Ruben Villegas, David Ha,
  Honglak Lee, and James Davidson.
\newblock Learning latent dynamics for planning from pixels.
\newblock In {\em ICML}, pages 2555--2565. PMLR, 2019.

\bibitem[\protect\citeauthoryear{He \bgroup \em et al.\egroup
  }{2020}]{he2020momentum}
Kaiming He, Haoqi Fan, Yuxin Wu, Saining Xie, and Ross Girshick.
\newblock Momentum contrast for unsupervised visual representation learning.
\newblock In {\em Proceedings of the IEEE/CVF Conference on Computer Vision and
  Pattern Recognition}, pages 9729--9738, 2020.

\bibitem[\protect\citeauthoryear{Hessel \bgroup \em et al.\egroup
  }{2018}]{hessel2018rainbow}
Matteo Hessel, Joseph Modayil, Hado Van~Hasselt, Tom Schaul, Georg Ostrovski,
  Will Dabney, Dan Horgan, Bilal Piot, Mohammad Azar, and David Silver.
\newblock Rainbow: Combining improvements in deep reinforcement learning.
\newblock In {\em AAAI}, 2018.

\bibitem[\protect\citeauthoryear{Kaiser \bgroup \em et al.\egroup
  }{2019}]{kaiser2019model}
Lukasz Kaiser, Mohammad Babaeizadeh, Piotr Milos, Blazej Osinski, Roy~H
  Campbell, Konrad Czechowski, Dumitru Erhan, Chelsea Finn, Piotr Kozakowski,
  Sergey Levine, et~al.
\newblock Model-based reinforcement learning for atari.
\newblock {\em arXiv preprint arXiv:1903.00374}, 2019.

\bibitem[\protect\citeauthoryear{Kielak}{2019}]{kielak2019recent}
Kacper~Piotr Kielak.
\newblock Do recent advancements in model-based deep reinforcement learning
  really improve data efficiency?
\newblock 2019.

\bibitem[\protect\citeauthoryear{Kingma and Ba}{2014}]{kingma2014adam}
Diederik~P Kingma and Jimmy Ba.
\newblock Adam: A method for stochastic optimization.
\newblock {\em arXiv preprint arXiv:1412.6980}, 2014.

\bibitem[\protect\citeauthoryear{Laskin \bgroup \em et al.\egroup
  }{2020a}]{laskin2020curl}
Michael Laskin, Aravind Srinivas, and Pieter Abbeel.
\newblock Curl: Contrastive unsupervised representations for reinforcement
  learning.
\newblock In {\em ICML}, pages 5639--5650. PMLR, 2020.

\bibitem[\protect\citeauthoryear{Laskin \bgroup \em et al.\egroup
  }{2020b}]{laskin2020reinforcement}
Misha Laskin, K.~Lee, A.~Stooke, L.~Pinto, Pieter Abbeel, and Aravind Srinivas.
\newblock Reinforcement learning with augmented data.
\newblock {\em NeurIPS}, 33, 2020.

\bibitem[\protect\citeauthoryear{Lee \bgroup \em et al.\egroup
  }{2019}]{lee2019stochastic}
Alex~X Lee, Anusha Nagabandi, Pieter Abbeel, and Sergey Levine.
\newblock Stochastic latent actor-critic: Deep reinforcement learning with a
  latent variable model.
\newblock {\em arXiv preprint arXiv:1907.00953}, 2019.

\bibitem[\protect\citeauthoryear{LEE \bgroup \em et al.\egroup
  }{2020}]{lee2020network}
KIMIN LEE, Kibok Lee, Jinwoo Shin, and Honglak Lee.
\newblock Network randomization: A simple technique for generalization in deep
  reinforcement learning.
\newblock In {\em Eighth International Conference on Learning Representations,
  ICLR 2020}. International Conference on Learning Representations, 2020.

\bibitem[\protect\citeauthoryear{Lei~Ba \bgroup \em et al.\egroup
  }{2016}]{lei2016layer}
Jimmy Lei~Ba, Jamie~Ryan Kiros, and Geoffrey~E Hinton.
\newblock Layer normalization.
\newblock {\em ArXiv e-prints}, pages arXiv--1607, 2016.

\bibitem[\protect\citeauthoryear{Li \bgroup \em et al.\egroup
  }{2021}]{li2021mural}
Kevin Li, Abhishek Gupta, Ashwin Reddy, Vitchyr~H Pong, Aurick Zhou, Justin Yu,
  and Sergey Levine.
\newblock Mural: Meta-learning uncertainty-aware rewards for outcome-driven
  reinforcement learning.
\newblock In {\em ICML}. PMLR, 2021.

\bibitem[\protect\citeauthoryear{Lin \bgroup \em et al.\egroup
  }{2019}]{lin2019towards}
Yijiong Lin, Jiancong Huang, Matthieu Zimmer, Juan Rojas, and Paul Weng.
\newblock Towards more sample efficiency in reinforcement learning with data
  augmentation.
\newblock {\em arXiv preprint arXiv:1910.09959}, 2019.

\bibitem[\protect\citeauthoryear{Liquin and
  Lombrozo}{2020}]{liquin2020explanation}
Emily~G Liquin and Tania Lombrozo.
\newblock Explanation-seeking curiosity in childhood.
\newblock {\em Current Opinion in Behavioral Sciences}, 35:14--20, 2020.

\bibitem[\protect\citeauthoryear{Malik \bgroup \em et al.\egroup
  }{2021}]{pmlr-v139-malik21c}
Dhruv Malik, Aldo Pacchiano, Vishwak Srinivasan, and Yuanzhi Li.
\newblock Sample efficient reinforcement learning in continuous state spaces: A
  perspective beyond linearity.
\newblock In {\em ICML}. PMLR, 2021.

\bibitem[\protect\citeauthoryear{Mnih \bgroup \em et al.\egroup
  }{2013}]{mnih2013playing}
Volodymyr Mnih, Koray Kavukcuoglu, David Silver, Alex Graves, Ioannis
  Antonoglou, Daan Wierstra, and Martin Riedmiller.
\newblock Playing atari with deep reinforcement learning.
\newblock {\em arXiv preprint arXiv:1312.5602}, 2013.

\bibitem[\protect\citeauthoryear{Mnih \bgroup \em et al.\egroup
  }{2015}]{mnih2015human}
Volodymyr Mnih, Koray Kavukcuoglu, David Silver, Andrei~A Rusu, Joel Veness,
  Marc~G Bellemare, Alex Graves, Martin Riedmiller, Andreas~K Fidjeland, Georg
  Ostrovski, et~al.
\newblock Human-level control through deep reinforcement learning.
\newblock {\em nature}, 518(7540):529--533, 2015.

\bibitem[\protect\citeauthoryear{Mnih \bgroup \em et al.\egroup
  }{2016}]{mnih2016asynchronous}
Volodymyr Mnih, Adria~Puigdomenech Badia, Mehdi Mirza, Alex Graves, Timothy
  Lillicrap, Tim Harley, David Silver, and Koray Kavukcuoglu.
\newblock Asynchronous methods for deep reinforcement learning.
\newblock In {\em ICML}. PMLR, 2016.

\bibitem[\protect\citeauthoryear{Nguyen \bgroup \em et al.\egroup
  }{2021}]{nguyen2021sample}
Thanh Nguyen, Tung~M Luu, Thang Vu, and Chang~D Yoo.
\newblock Sample-efficient reinforcement learning representation learning with
  curiosity contrastive forward dynamics model.
\newblock {\em arXiv preprint arXiv:2103.08255}, 2021.

\bibitem[\protect\citeauthoryear{Oord \bgroup \em et al.\egroup
  }{2018}]{oord2018representation}
Aaron van~den Oord, Yazhe Li, and Oriol Vinyals.
\newblock Representation learning with contrastive predictive coding.
\newblock {\em arXiv preprint arXiv:1807.03748}, 2018.

\bibitem[\protect\citeauthoryear{Paszke \bgroup \em et al.\egroup
  }{2019}]{paszke2019pytorch}
Adam Paszke, Sam Gross, Francisco Massa, Adam Lerer, James Bradbury, Gregory
  Chanan, Trevor Killeen, Zeming Lin, Natalia Gimelshein, Luca Antiga, et~al.
\newblock Pytorch: An imperative style, high-performance deep learning library.
\newblock {\em Advances in neural information processing systems},
  32:8026--8037, 2019.

\bibitem[\protect\citeauthoryear{Pathak \bgroup \em et al.\egroup
  }{2017}]{pathak2017curiosity}
Deepak Pathak, Pulkit Agrawal, Alexei~A Efros, and Trevor Darrell.
\newblock Curiosity-driven exploration by self-supervised prediction.
\newblock In {\em ICML}, pages 2778--2787. PMLR, 2017.

\bibitem[\protect\citeauthoryear{Raileanu \bgroup \em et al.\egroup
  }{2020}]{raileanu2020automatic}
Roberta Raileanu, Max Goldstein, Denis Yarats, Ilya Kostrikov, and Rob Fergus.
\newblock Automatic data augmentation for generalization in deep reinforcement
  learning.
\newblock {\em arXiv preprint arXiv:2006.12862}, 2020.

\bibitem[\protect\citeauthoryear{Rakelly \bgroup \em et al.\egroup
  }{2019}]{rakelly2019efficient}
Kate Rakelly, Aurick Zhou, Chelsea Finn, Sergey Levine, and Deirdre Quillen.
\newblock Efficient off-policy meta-reinforcement learning via probabilistic
  context variables.
\newblock In {\em ICML}, pages 5331--5340. PMLR, 2019.

\bibitem[\protect\citeauthoryear{Saxe \bgroup \em et al.\egroup
  }{2013}]{saxe2013exact}
Andrew~M Saxe, James~L McClelland, and Surya Ganguli.
\newblock Exact solutions to the nonlinear dynamics of learning in deep linear
  neural networks.
\newblock {\em arXiv preprint arXiv:1312.6120}, 2013.

\bibitem[\protect\citeauthoryear{Schaul \bgroup \em et al.\egroup
  }{2015}]{schaul2015prioritized}
Tom Schaul, John Quan, Ioannis Antonoglou, and David Silver.
\newblock Prioritized experience replay.
\newblock {\em arXiv preprint arXiv:1511.05952}, 2015.

\bibitem[\protect\citeauthoryear{Schaul \bgroup \em et al.\egroup
  }{2016}]{schaul2016prioritized}
Tom Schaul, John Quan, Ioannis Antonoglou, and David Silver.
\newblock Prioritized experience replay.
\newblock In {\em ICLR}, 2016.

\bibitem[\protect\citeauthoryear{Schwarzer \bgroup \em et al.\egroup
  }{2020}]{schwarzer2020data}
Max Schwarzer, Ankesh Anand, Rishab Goel, R~Devon Hjelm, Aaron Courville, and
  Philip Bachman.
\newblock Data-efficient reinforcement learning with self-predictive
  representations.
\newblock In {\em ICLR}, 2020.

\bibitem[\protect\citeauthoryear{Seo \bgroup \em et al.\egroup
  }{2021}]{seo_chen2021re3}
Younggyo Seo, Lili Chen, Jinwoo Shin, Honglak Lee, Pieter Abbeel, and Kimin
  Lee.
\newblock State entropy maximization with random encoders for efficient
  exploration.
\newblock {\em arXiv preprint arXiv:2102.09430}, 2021.

\bibitem[\protect\citeauthoryear{Spielberger and
  Starr}{2012}]{spielberger2012curiosity}
Charles~D Spielberger and Laura~M Starr.
\newblock Curiosity and exploratory behavior.
\newblock In {\em Motivation: Theory and research}, pages 231--254. 2012.

\bibitem[\protect\citeauthoryear{Sun \bgroup \em et al.\egroup
  }{2022}]{sun2022psychological}
Chenyu Sun, Hangwei Qian, and Chunyan Miao.
\newblock From psychological curiosity to artificial curiosity:
  Curiosity-driven learning in artificial intelligence tasks.
\newblock {\em arXiv preprint arXiv:2201.08300}, 2022.

\bibitem[\protect\citeauthoryear{Sutton and
  Barto}{1998}]{sutton1998reinforcement}
Richard~S Sutton and Andrew~G Barto.
\newblock Reinforcement learning: an introduction mit press.
\newblock {\em Cambridge, MA}, 22447, 1998.

\bibitem[\protect\citeauthoryear{Tieleman \bgroup \em et al.\egroup
  }{2012}]{tieleman2012lecture}
Tijmen Tieleman, Geoffrey Hinton, et~al.
\newblock Lecture 6.5-rmsprop: Divide the gradient by a running average of its
  recent magnitude.
\newblock {\em COURSERA: Neural networks for machine learning}, 4(2):26--31,
  2012.

\bibitem[\protect\citeauthoryear{Todorov \bgroup \em et al.\egroup
  }{2012}]{todorov2012mujoco}
Emanuel Todorov, Tom Erez, and Yuval Tassa.
\newblock Mujoco: A physics engine for model-based control.
\newblock In {\em 2012 IEEE/RSJ International Conference on Intelligent Robots
  and Systems}, pages 5026--5033. IEEE, 2012.

\bibitem[\protect\citeauthoryear{Tunyasuvunakool \bgroup \em et al.\egroup
  }{2020}]{tunyasuvunakool2020dm_control}
Saran Tunyasuvunakool, Alistair Muldal, Yotam Doron, Siqi Liu, Steven Bohez,
  Josh Merel, Tom Erez, Timothy Lillicrap, Nicolas Heess, and Yuval Tassa.
\newblock dm\_control: Software and tasks for continuous control.
\newblock {\em Software Impacts}, 6:100022, 2020.

\bibitem[\protect\citeauthoryear{Van~Hasselt \bgroup \em et al.\egroup
  }{2016}]{van2016deep}
Hado Van~Hasselt, Arthur Guez, and David Silver.
\newblock Deep reinforcement learning with double q-learning.
\newblock In {\em Proceedings of the AAAI conference on artificial
  intelligence}, volume~30, 2016.

\bibitem[\protect\citeauthoryear{van Hasselt \bgroup \em et al.\egroup
  }{2019}]{van2019use}
Hado~P van Hasselt, Matteo Hessel, and John Aslanides.
\newblock When to use parametric models in reinforcement learning?
\newblock {\em NeurIPS}, 32:14322--14333, 2019.

\bibitem[\protect\citeauthoryear{Wang \bgroup \em et al.\egroup
  }{2016}]{wang2016dueling}
Ziyu Wang, Tom Schaul, Matteo Hessel, Hado Hasselt, Marc Lanctot, and Nando
  Freitas.
\newblock Dueling network architectures for deep reinforcement learning.
\newblock In {\em International conference on machine learning}, pages
  1995--2003. PMLR, 2016.

\bibitem[\protect\citeauthoryear{Wang \bgroup \em et al.\egroup
  }{2020}]{wang2020improving}
Kaixin Wang, Bingyi K, Jie S, and Jiashi F.
\newblock Improving generalization in reinforcement learning with mixture
  regularization.
\newblock In {\em NeurIPS}, 2020.

\bibitem[\protect\citeauthoryear{Yarats \bgroup \em et al.\egroup
  }{2020}]{yarats2020image}
Denis Yarats, Ilya Kostrikov, and Rob Fergus.
\newblock Image augmentation is all you need: Regularizing deep reinforcement
  learning from pixels.
\newblock In {\em ICLR}, 2020.

\bibitem[\protect\citeauthoryear{Yarats \bgroup \em et al.\egroup
  }{2021a}]{yarats2021drqv2}
Denis Yarats, Rob Fergus, Alessandro Lazaric, and Lerrel Pinto.
\newblock Mastering visual continuous control: Improved data-augmented
  reinforcement learning.
\newblock {\em arXiv preprint arXiv:2107.09645}, 2021.

\bibitem[\protect\citeauthoryear{Yarats \bgroup \em et al.\egroup
  }{2021b}]{yarats2021improving}
Denis Yarats, Amy Zhang, Ilya Kostrikov, Brandon Amos, Joelle Pineau, and Rob
  Fergus.
\newblock Improving sample efficiency in model-free reinforcement learning from
  images.
\newblock In {\em AAAI}, 2021.

\bibitem[\protect\citeauthoryear{Zhao and Tresp}{2019}]{zhao2019curiosity}
Rui Zhao and Volker Tresp.
\newblock Curiosity-driven experience prioritization via density estimation.
\newblock {\em arXiv preprint arXiv:1902.08039}, 2019.

\end{thebibliography}

\normalsize
\clearpage
\appendixtitleon
\appendixtitletocon
\begin{appendices}
\section{Extended Background}
\label{app:extended_background}
\subsection{Efficiency of Data Augmentation in RL}
Data augmentation such as translation, crop, rotation, and cutout has been widely applied in computer vision but is only recently introduced in RL to incorporate invariances for representation learning \cite{cobbe2019quantifying,lin2019towards,raileanu2020automatic,lee2020network,laskin2020reinforcement,yarats2020image,laskin2020curl}. In particular, RAD \cite{laskin2020reinforcement} investigates several commonly used data augmentation methods, where random crop and translation are found most effective. Meanwhile, DrQ \cite{yarats2020image} regularizes the Q-function to ensure that multiple augmentations from the same observation should have similar Q-values. CURL \cite{laskin2020curl} leverages contrastive learning with data augmentation such that representation learning becomes more robust. When learning directly from raw pixels, these works have outperformed other methods, including pixel SAC \cite{haarnoja2018soft}, PlaNet \cite{hafner2019learning}, Dreamer \cite{hafner2019dream}, SLAC \cite{lee2019stochastic} and SAC+AE \cite{yarats2021improving}. However, sample importance has not been fully exploited in any data augmentation based RL.

\subsection{Soft Actor-Critic (SAC)}
We consider a Markov Decision Process (MDP) setting with the state $s_t \in \mathcal{S}$, the action $a_t \in \mathcal{A}$, the transition probability $P$ mapping from the current state $s_t$ and action $a_t$ to the next state $s'_{t}$, and the (extrinsic) reward $r^e_t \in \mathcal{R}$. SAC \cite{haarnoja2018soft} is an off-policy model-free reinforcement learning (RL) algorithm that learns a stochastic policy $\pi_\psi$ (actor) with state-action value functions $Q_{\phi_1}, Q_{\phi_2}$ (critics), and a temperature coefficient $\alpha$ by encouraging exploration through a $\gamma$-discounted maximum-policy-entropy term. However, agents are often required to learn directly from high-dimensional visual observations $o_t \in \mathcal{O}$ rather than states $s_t$ in practice. 

In particular, the critics are trained by minimizing the squared Bellman error via uniformly sampling transitions $\tau_t = (o_t, a_t, r^e_t, d_t, o'_{t})$ from a replay buffer $\mathcal{B}$, 
\begin{equation*}
\label{SAC_critic_loss}
\mathcal{L}_Q(\phi)= \mathbb{E}_{\tau \sim \mathcal{B}} \left[ \left( Q_\phi (o_t, a_t) - (r^e_t+\gamma (1-d_t)\mathcal{T}_t)\right)^2 \right]
\end{equation*}
where $d_t$ is the done signal and the soft target value $\mathcal{T}_t$ is implicitly parameterized as
\begin{equation*}
\mathcal{T}_t=\min_{i=1,2}Q_{\bar{\phi_i}}(o'_{t},a')-\alpha \log \pi_\psi (a' | o'_{t}).
\end{equation*}
$Q_{\bar{\phi_i}}$ is the exponential moving average (EMA) of $Q_{\phi_i}$ to impose training stability, and $a'$ is sampled stochastically by the actor network using the next observation $o'_t$. 

The actor network $\pi_\psi (a|o)$ is a parametric tanh-Gaussian which stochastically samples $a=\tanh (\mu_\psi(o)+\sigma_\psi(o)\epsilon)$ given the observation $o$, where $\epsilon \sim \mathcal{N}(0,1)$, and 
$\mu_\psi$ and $\sigma_\psi$ are the parametric mean and standard deviation, respectively. Meanwhile, the actor network is trained by minimizing the divergence from the exponential of the soft-Q function, equivalent to
\begin{equation*}
\mathcal{L}_\pi(\psi)=-\mathbb{E}_{a\sim \pi}\left[\min_{i=1,2}Q_{\phi_i}(o_{t},a)-\alpha \log \pi_\psi (a | o_{t})\right].
\end{equation*}

Finally, $\alpha$ controls the priority of entropy maximization and is optimized against a target entropy as
\begin{equation*}
\mathcal{L}(\alpha)=\mathbb{E}_{a\sim \pi} \left[ -\alpha \log \pi_\psi (a | o_{t}) - \alpha \bar{\mathcal{H}}\right]
\end{equation*}
where $\bar{\mathcal{H}} \in \mathbb{R}$ represents the target entropy with $\bar{\mathcal{H}}=-|\mathcal{A}|$.

\subsection{Rainbow DQN}

Rainbow DQN \cite{hessel2018rainbow} is an off-policy deep RL algorithm for discrete action space, extending the conventional DQN \cite{mnih2015human} with multiple improvements of other DQN methods into a single learner. In particular, it utilizes the Double Q-Learning method \cite{van2016deep} to resolve overestimation bias. As an off-policy algorithm, it employs the Prioritized Experience Replay technique \cite{schaul2015prioritized} to sample the novel transitions with high TD-errors more frequently. In addition, the dueling critic architecture network \cite{wang2016dueling} is incorporated to learn valuable states, avoiding determining the effect of all actions and states. Instead of the expected return, it uses distributional reinforcement learning to output a distribution over possible value function bins, with $n$-step learning \cite{sutton1998reinforcement} and noisy linear layers \cite{fortunato2017noisy} for better exploration. Combining all above-mentioned techniques and architectures, Rainbow DQN can obtain state-of-the-art sample efficiency on Atari Games \cite{bellemare2013arcade} benchmarks. To further enhance the sample efficiency, an optimized configuration of Rainbow DQN hyperparameters is proposed to 
benchmark Atari Games at 100K training steps \cite{van2019use}.

\subsection{Advantage Actor-Critic (A2C)}
In contrast to SAC and Rainbow DQN from off-policy RL algorithms, Advantage Actor Critic (A2C) \cite{mnih2016asynchronous} is an on-policy algorithm that combines the policy-based method (an actor) and value-based method (a critic), with an advantage function. While the value function measures how good the agent is at each state, the advantage function $A(s, a)$ captures how better an action $a$ against the others at a given state $s$,
\begin{equation*}
A(s, a)= Q(s,a)-V(s).
\end{equation*}
Subsequently, the actor can leverage this information to optimize the agent's policy. As the result, the high variance of policy gradient can be greatly reduced to stabilize the model. Based on A2C, RE3 \cite{seo_chen2021re3} proposed to incorporate the state entropy as an intrinsic reward to encourage efficient exploration, which can be estimated with a fixed random encoder. RE3 has been shown to be capable of improving the sample efficiency on navigation tasks from the MiniGrid \cite{gym_minigrid} benchmark.

\subsection{Contrastive Unsupervised RL (CURL)}
To learn a meaningful representation from raw pixels, CURL \cite{laskin2020curl} utilizes data augmentation and contrastive learning to train an image encoder $f_\theta(o)$ in a self-supervised way, imposing an instance discrimination between similar and dissimilar encoded states. Given a batch of visual observations $o$, each is augmented twice and encoded into a query $q=f_\theta(o_q)$ and a key $k=f_{\bar{\theta}} (o_k)$, where the positive keys $k^+$ indicate that both $o_k$ and $o_q$ are augmented from the same $o$ and otherwise negative keys $k^{-}$. The key encoder $f_{\bar{\theta}}$ is a momentum-based moving average \cite{he2020momentum} of the query encoder $f_\theta$ to ensure consistency and stability, where $f_\theta$ is learned by enforcing $q$ to match with $k^+$ and keep far apart from $k^-$. The contrastive loss term to optimize the query encoder is defined as
\begin{equation*}
\mathcal{L}_f(\theta) = -\sum_{b=1}^{B}\log 
\frac{\text{sim}(q_b, k^+)}{\text{sim}(q_b, k^+)+\sum_{l=1}^{B-1}\text{sim} (q_b, k_l^-)}
\end{equation*}
where $B$ is the batch size and $\text{sim} (q, k)=q^TWk$ is the bi-linear inner-product to measure the similarity between $q$ and $k$ \cite{oord2018representation}. 

\section{Experimental Settings}
We conduct experiments on four cloud servers and one physical server with the following configurations.
\begin{itemize}
	\item Operation System: Ubuntu 18.04
	\item Memory: 32GiB / 32GiB / 32GiB / 32GiB / 128GiB
	\item CPU: Intel Core Processor (Skylake) / Intel Core Processor (Skylake) / Intel Core Processor (Skylake) / Intel Core Processor (Skylake) / Intel(R) Xeon(R) CPU E5-2620 v2 @ 2.10GHz
	\item vCPU: 8 / 8 / 16 / 16 / 24
	\item GPU: 2 $\times$ NVIDIA Tesla P100 16GB / 2 $\times$ NVIDIA Tesla P100 16GB / 1 $\times$ NVIDIA Tesla V100S PCIE 32GB / 1 $\times$ NVIDIA Tesla V100S PCIE 32GB / 2 $\times$ NVIDIA GeForce RTX 3090 24GB
\end{itemize}
Our proposed CCLF is implemented using PyTorch \cite{paszke2019pytorch} based on CURL \cite{laskin2020curl}, where the contrastive learning module is incorporated to jointly improve the representation learning and Q-learning. 
As CCLF is model-agnostic, it can be applied to different base RL algorithms with some adaptions. In this paper, we mainly focus on SAC \cite{haarnoja2018soft} as the base algorithm while additional experiments with Rainbow DQN \cite{hessel2018rainbow} and A2C \cite{mnih2016asynchronous} have also been carried out on a diverse range of exploration tasks from Atari Games \cite{bellemare2013arcade} and MiniGrid \cite{gym_minigrid}. The implementation details are as follows.

\subsection{Implementation of CCLF on SAC for Continuous Control Tasks in DMC Suite}
\label{app:SAC_settings}
\subsubsection{Encoder Network}
We utilize the same encoder network as SAC-AE \cite{yarats2021improving}. In particular, it is parameterized as 4 convolution layers with the ReLU activation, $3 \times 3$ kernels, and 32 channels. The stride is set to be 2 for the first layer and is reduced to 1 thereafter. Subsequently, the output is passed to a single normalized fully-connected layer by LayerNorm \cite{lei2016layer} with the tanh nonlinearity applied at the end. Finally, the orthogonal initialization \cite{saxe2013exact} is performed to initialize the parameters for all convolutional layers as well as the fully-connected linear layer, where the bias is set to be zero.

\subsubsection{Actor and Critic Networks}
Both the actor and critic networks employ the image encoder network as described above to encode pixel observations, and their convolution layers share the same weights. During batch learning, these weights can only be updated by the critic optimizer and the contrastive optimizer. In other words, we detach the actor encoder to stop the propagation of the shared convolution layers during the actor update. 

For the actor network, the encoded state representation is fed into a 3-layer MLP with the ReLU activation to output the policy mean and covariance for the diagonal Gaussian. Similarly, the critic networks pass the encoded state representation into a 3-layer MLP with the ReLU activation after each layer except for the last one. For both actor and critic networks, we set the hidden dimension to 1024.

It should be noted that the clipped double Q-learning \cite{van2016deep} is utilized for critic networks, a popular technique for stability in SAC-based methods. Meanwhile, the target critic network is momentum-based updated by the critic network to impose Q-learning stability. When we train the critics, only the parameters of the critic networks are updated by the optimizer. 

\subsubsection{Contrastive Learning}
Following the same implementation of CURL \cite{laskin2020curl}, the contrastive learning module is built with the image encoder networks in the critics. In particular, the query is encoded by the encoder in the critic network and the keys are encoded by the encoder in the target network. The key encoder is also momentum-based updated by the query encoder to avoid instability \cite{he2020momentum}. To measure the similarity between query and keys, the bi-linear inner-product is utilized and its weight is initialized randomly.

\subsubsection{Training and Evaluation Setup}
During the model training, the 100 $\times$ 100 visual observations are randomly cropped to 84 $\times$ 84 pixels. We first collect 1000 transitions using a random policy and store them in the replay buffer. After that, the subsequent transition tuples are collected by sampling actions from the current policy. Meanwhile, the model update is performed by sampling a mini-batch of transition tuples from the replay buffer at each training step. For clarification, the environment step in this section refers to the multiplication of action repeat and training steps. For example, 50K training steps with the action repeat of 2 represents 100K environment steps. For a fair evaluation, all models are trained with 6 random seeds, during which agents are evaluated every 10K environment steps with 10 episodes. 
During the evaluation, the 100 $\times$ 100 visual observations are center cropped to 84 $\times$ 84 pixels and agents perform with the deterministic policy outputted by the actor network instead of sampling a stochastic action.

\subsubsection{Environments}
We evaluate the proposed CCLF from the perspectives of both sample efficiency and ultimate performance on six continuous control tasks in the DMC suite~\cite{tunyasuvunakool2020dm_control}. The DMC suite provides a diverse collection of complex control using MuJoCo physics \cite{todorov2012mujoco} with different settings such as sparse rewards and complex dynamics, which are challenging for agents to operate directly from visual observations. The six tasks are commonly used as benchmarks to evaluate the agent's performance, where the evaluation metric is the performance score in the range of [0, 1000].

\subsubsection{Hyper-parameters}
Throughout all experiments, the network hyper-parameters are kept fixed and consistent unless otherwise stated. We employ the Adam optimizer \cite{kingma2014adam} and set the batch size to 512. To capture both spatial and temporal information, 3 consecutive frames are stacked into one observation. The detailed setting of hyper-parameters for baseline models is listed in Table \ref{common_settings}, which follows the same settings of CURL \cite{laskin2020curl} and DrQ \cite{yarats2020image}. 

\begin{table}[h!]
	\begin{tabular}{@{}ll@{}}
		\toprule
		Hyper-parameters                 & Value                        \\ \midrule
		Observation rendering          & (100, 100)                   \\
		Observation downsampling       & (84, 84)                     \\
		Replay buffer size             & 100K                     \\
		Initial steps                  & 1K                         \\
		Stacked frames                 & 3                            \\
		Action repeat                  & 2 Finger, Spin; Walker, walk \\
		& 8 Cartpole, Swingup          \\
		& 4 otherwise                  \\
		Hidden units (MLP)             & 1024                         \\
		Evaluation episodes            & 10                           \\
		Optimizer                      & Adam                         \\
		($\beta1, \beta2$) $\to$ ($f_\theta$, $\pi_\psi$, $Q_\phi$)        & (.9, .999)                   \\
		($\beta1, \beta2$) $\to$ $\alpha$                 & (.5, .999)                   \\
		Learning rate ($f_\theta$, $\pi_\psi$, $Q_\phi$)      & $2e-4$ cheetah, run          \\
		& $1e-3$ otherwise             \\
		Learning rate              & $1e-4$                       \\
		Batch Size                     & 512                          \\
		Initial temperature            & 0.1                          \\
		Q function EMA rate              & 0.01                         \\
		Critic target update frequency & 2                            \\
		Actor update frequency         & 2                            \\
		Actor log stddev bounds        & {[}-10, 2{]}                 \\
		Convolutional layers           & 4                            \\
		Number of filters              & 32                           \\
		Activation                     & ReLU                         \\
		Encoder EMA rate                   & 0.05                         \\
		Encoder feature dimension      & 50                           \\
		Discount factor                      & 0.99                         \\ \bottomrule
	\end{tabular}
	\caption{Hyper-parameters used for CURL and DrQ based models with SAC being the base algorithm. Most hyper-parameters are fixed for all tasks while only action repeat and learning rate vary across different tasks.}
	\label{common_settings}
\end{table}

For the proposed CCLF, we initialize it with $[K, M]=[5,5]$ to generate a sufficiently large amount of augmented inputs. In addition to the hyper-parameters set in Table \ref{common_settings}, we list the other hyper-parameters used in CCLF in Table \ref{curiosity_setting}, which are fixed when evaluating on the six continuous control tasks in the DMC suite.

\begin{table}[h!]
	\begin{tabular}{p{6cm}p{1.5cm}} 
		\toprule
		Hyper-parameters                 & Value                        \\ \midrule
		$[K, M]$ & $[5, 5]$ \\
		Momentum coefficient for prioritization & 0.99 \\
		Intrinsic reward decay weight & 2e-5 \\ 
		Intrinsic reward temperature  & 0.2 \\
		\bottomrule
	\end{tabular}
	\caption{Additional hyper-parameters used for the proposed CCLF on SAC, fixed across all evaluated tasks.}
	\label{curiosity_setting}
\end{table}

\subsection{Implementation of CCLF on Rainbow DQN for Discrete Control Tasks in Atari Games}
\label{app:rainbow}
Following the similar setting and rationale in CURL, we are able to extend CCLF on discrete control tasks from Atari100K with minimal modifications. In particular, we select the Rainbow DQN with the data-efficient architecture (Efficient Rainbow) \cite{van2019use} as the base algorithm for our proposed model-agnostic CCLF. For the encoder network, actor and critic networks, and contrastive learning module, we follow the exactly same network architectures as CURL Rainbow, from the publicly available repository \url{https://github.com/aravindsrinivas/curl_rainbow}. To incorporate the curiosity-driven experienced replay component, we modify the algorithm by replacing the TD-error-based prioritization with our proposed contrastive-curiosity-based prioritization, where the sampling weights are updated in a momentum manner.
\subsubsection{Training and Evaluation Setup}
During the model training, the 84 $\times$ 84 visual observations are randomly cropped to 80 $\times$ 80 pixels, padded by 4 pixels and then randomly cropped again to 84 $\times$ 84 pixels. We first collect 1600 transitions using a random policy and store them in the replay buffer. After that, the subsequent transition tuples are collected using the outputted actions from the current policy. Meanwhile, the model update is performed by sampling a mini-batch of transition tuples from the replay buffer at each training step. 
For a fair evaluation, our CCLF is trained with 4 random seeds, during which agents are evaluated every 10K training steps with 10 episodes. 
During the evaluation, the 84 $\times$ 84 rendered observations are directly inputted into the trained model without further augmentations. We evaluate CCLF from the perspective of sample efficiency at 100K training steps on 26 discrete control tasks in the Atari Games.

\subsubsection{Hyper-parameters}
Throughout all experiments, the network hyper-parameters are kept fixed and consistent unless otherwise stated. For a fair comparison and evaluation, the same hyper-parameters used in CURL on the Rainbow DQN are also utilized in our experiments. To further stabilize the model training, we set a coefficient for the contrastive loss term, which is consistent with the CURL method. We employ the Adam optimizer \cite{kingma2014adam} and set the batch size to 32. To capture both spatial and temporal information, 4 consecutive frames are stacked into one observation. The detailed setting of hyper-parameters is provided in Table~\ref{rainbow_setting} and Table~\ref{rainbow_setting_CCLF}. 

\begin{table}[h!]
	\begin{tabular}{p{4cm}p{3.5cm}} 
		\toprule
		Hyperparameter                            & Value                   \\ \midrule
		Data   augmentation              & Random Crop                                         \\
		Grey-scaling                              & TRUE                    \\
		Observation                               & (84, 84)                \\
		Frames stacked                            & 4                       \\
		Action repetitions                        & 4                       \\
		Replay buffer size                        & 100K                    \\
		Replay period                             & 1                       \\
		Min replay size for sampling              & 1600                    \\
		Training steps                            & 100K                    \\
		Training frames                           & 400K                    \\
		Reward clipping                           & {[}-1,1{]}              \\
		Max frames per episode                    & 108K                    \\
		Minibatch size                            & 32                      \\
		Discount factor                           & 0.99                    \\
		Optimizer                                 & Adam                    \\
		Optimizer: learning rate                  & 0.0001                  \\
		Optimizer: $\beta_1$                      & 0.9                     \\
		Optimizer: $\beta_2$                      & 0.999                   \\
		Optimizer: $\epsilon$                     & 0.00015                 \\
		Max gradient norm                         & 10                      \\
		Q network: channels                       & 32, 64                  \\
		Q network: filter size                    & 5 $\times$ 5, 5 $\times$ 5            \\
		Q network: stride                         & 5, 5             \\
		Q network: hidden units                   & 256                     \\
		Non-linearity                             & ReLU                    \\
		Multi step return                         & 20                      \\
		Update                                    & Distributional Double Q \\
		Support-of-Q-distribution                 & 51 bins                 \\
		Dueling                                   & TRUE                    \\
		Target network update period              & Every 2000 updates      \\
		Exploration                               & Noisy Nets              \\
		Noisy nets parameter 0.1                  & 0.1                     \\
		Momentum (EMA for key encoder   update)   & 0.001               \\* \bottomrule
	\end{tabular}
	\caption{Hyper-parameters used for CURL and DrQ based models with Rainbow DQN being the base algorithm. }
	\label{rainbow_setting}
\end{table}

\begin{table}[h!]
	\begin{tabular}{p{0.21\textwidth}p{0.22\textwidth}} 
		\toprule
		Hyperparameter                            & Value                   \\ \midrule
		$[K,M]$                                     & [5, 5]                    \\
		Coefficient of contrastive loss term& 1 for alien, bank heist, battle zone, kangaroo, kung fu master, and ms pacman; 0.0001 elsewise \\
		Momentum coefficient for   prioritization & 0.99                    \\
		Intrinsic reward decay weight             & 2e-5                    \\
		Intrinsic reward temperature              & 2e-4                    \\* \bottomrule
	\end{tabular}
	\caption{Additional hyper-parameters used for CCLF on Rainbow DQN. Most hyper-parameters are fixed for all tasks while only the coefficient of contrastive loss term varies across different tasks.}
	\label{rainbow_setting_CCLF}
\end{table}

\subsection{Implementation of CCLF on A2C for Navigation Tasks in MiniGrid}
\label{app:A2C}
In addition to the off-policy RL algorithms, we further demonstrate that CCLF can be applied to the on-policy RL algorithm A2C with minimal modifications. 
Note that A2C does not employ the replay buffer for experience replay, thus we remove the curiosity-based prioritization component. 
Moreover, we also apply our CCLF to RE3 that proposed another type of intrinsic reward to encourage entropy-based exploration in the base algorithm of A2C. In particular,
for the encoder network as well as actor and critic networks, we utilize the publicly available implementation repository \url{https://github.com/younggyoseo/RE3/tree/master/a2c_re3} with the default hyperparameters for the A2C implementation in MiniGrid benchmark.
For the contrastive learning module, the query network is built with the encoder network shared by the actor and critic, while the key network is momentum-based updated from the query. To measure the similarity between query and keys, the bi-linear inner-product is utilized and its weight is initialized randomly.


\subsubsection{Training and Evaluation Setup}
\begin{table}[h]
	\begin{tabular}{p{3.9cm}p{3.8cm}} 
		\toprule
		Hyperparameter                            & Value                   \\ \midrule
		Input Size                       & (7, 7, 3) \\
		Replay buffer size (for RE3 intrinsic   reward) & 10K                                                         \\
		Stacked frames                   & 1         \\
		Action repeat                    & 1         \\
		Evaluation episodes              & 100       \\
		Optimizer                        & RMSprop   \\
		Number of processes              & 16        \\
		Frames per process               & 8         \\
		Discount                         & 0.99      \\
		GAE $\lambda$                     & 0.95      \\
		Entropy coefficient              & 0.001     \\
		Value loss term coefficient      & 0.5       \\
		Maximum norm of gradient         & 0.5       \\
		RMSprop $\epsilon$               & 0.05      \\
		Clipping $\epsilon$              & 0.2       \\
		Recurrence                       & None      \\
		Training Steps & 400K for Empty-16x16 and DoorKey-6x6; 2400K for DoorKey-8x8 \\
		Evaluation frequency                            & 3200 for Empty-16x16 and DoorKey-6x6; 19200 for DoorKey-8x8     \\
		RE3: intrinsic reward coefficient               & 0.1 in Empty-16x16; 0.005 in DoorKey-6x6;001 in DoorKey-8x8 \\
		RE3: $k$                         & 3                  \\* \bottomrule
	\end{tabular}
	\caption{Hyper-parameters used for baselines of A2C and RE3. Most hyper-parameters are fixed for all tasks while the training steps,  evaluation frequency and RE3 intrinsic reward coefficient change across different tasks as specified in RE3 settings.}
	\label{A2C_settings}
\end{table}

During the model training, the $7\times7\times3$ input from MiniGrid is directly passed to the base RL learner and contrastive learning without augmentations or input selection. It is because any small augmentation on embedded values will cause highly inconsistent learned features. The on-policy agent first collects 8 frames of transitions per process with a total of 16 processes during one update.
For a fair evaluation, our CCLF is trained with 5 random seeds and the mean scores are plotted to demonstrate the learning capabilities. 
We evaluate the proposed CCLF from the perspective of sample efficiency at 400K training steps on Empty-16$\times$16 and DoorKey-6$\times$6 while 2400K steps on DoorKey-8$\times$8 in the MiniGrid, consistent with the baseline settings.

\subsubsection{Hyper-parameters}
In order to carefully examine the benefits of our CCLF without any other changes, we employ the exact same hyper-parameters specified in the RE3 paper appendix Section C \cite{seo_chen2021re3}. Throughout all experiments, the network hyper-parameters are kept fixed and consistent unless otherwise stated. To further stabilize the model training, we set a coefficient for the contrastive loss term. We employ the RMSprop optimizer \cite{tieleman2012lecture} and present a full list of hyperparameters that are used for baselines in Table~\ref{A2C_settings} as well as CCLF in Table~\ref{A2C_settings_CCLF}. 

\begin{table}[h]
	\begin{tabular}{p{6cm}p{1.8cm}} 
		\toprule
		Hyperparameter                            & Value                   \\ \midrule
		Contrastive: hidden units        & 128       \\
		Non-linearity                    & ReLU      \\
		Momentum (EMA for key encoder   update)   & 0.001  \\
		Coefficient for contrastive loss & 0.0001    \\
		Intrinsic reward decay weight    & 2e-5  \\
		Intrinsic reward temperature     & 2e-4                  \\* \bottomrule
	\end{tabular}
	\caption{Additional hyper-parameters used for the proposed CCLF on A2C and RE3, fixed across different tasks.}
	\label{A2C_settings_CCLF}
\end{table}

\section{Experimental Results}
\subsection{Additional Results on the DMC Suite}
\subsubsection{Overall Performance}

\begin{figure*}[!h]
	\centering
	\includegraphics[width=0.95\textwidth]{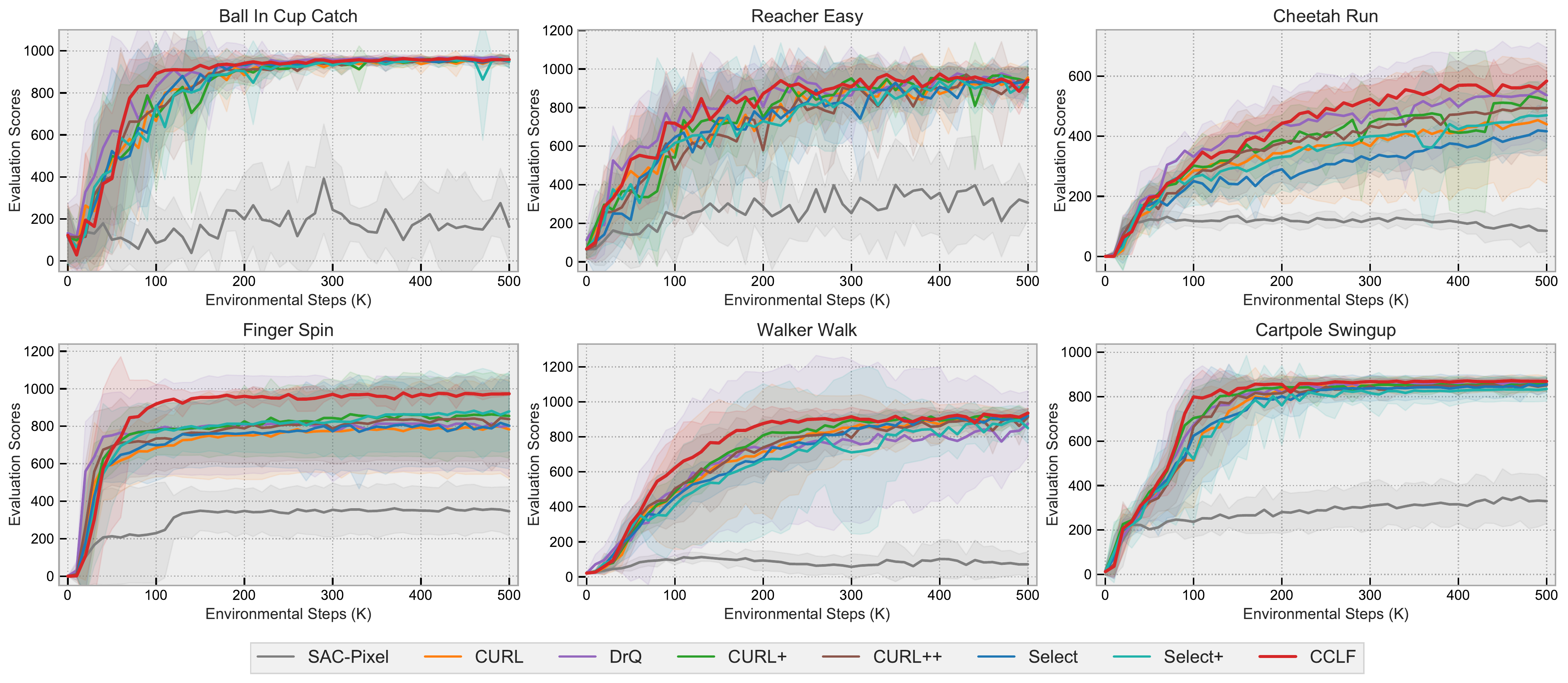} 
	\caption{Comparison of the sample efficiency and learning performances against baselines on the 6 continuous control tasks from the DMC Suite. Our proposed CCLF outperforms the other methods in terms of sample efficiency and converges much faster than the baselines.}
	\label{fig:SACresults}
\end{figure*}

\begin{figure*}[h!]
	\centering
	\includegraphics[width=0.9\textwidth]{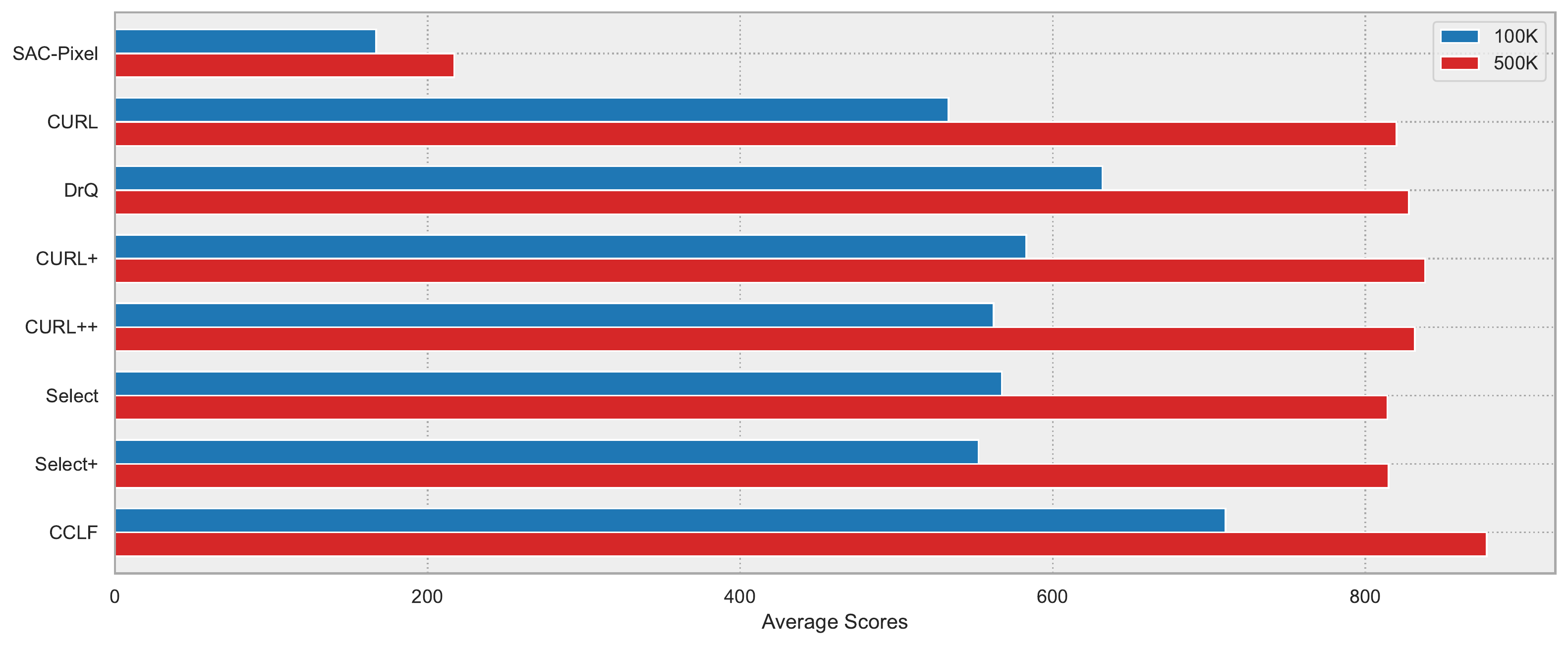} 
	\caption{Comparison of average scores at 100K and 500K environment scores for DMC Suite}
	\label{average_score}
\end{figure*}

The full learning performances on all six continuous control tasks are shown in Figure~\ref{fig:SACresults}, where the proposed CCLF enables the agent to outperform other baseline models in sample efficiency. To compare the overall performance of CCLF against the baselines, we average the performances over the 6 selected control tasks in the DMC suite. The results are shown in Figure~\ref{average_score}, where CCLF significantly surpasses the other models at both 100K and 500K environment steps. At 100K environment steps, our CCLF achieves a 1.14 $\times$ mean score of the second-best model (DrQ), indicating a significant improvement of the sample efficiency. At 500K environment steps, the average score of CCLF is 1.04 $\times$ mean score of the second-best model (CURL+). In a nutshell, our proposed CCLF can obtain state-of-the-art performance from the perspectives of sample efficiency and learning capabilities.

\subsubsection{Comparison of Computational Complexity}
\label{app:computational_cost}
\begin{figure*}[h!]
	\centering
	\includegraphics[width=0.9\textwidth]{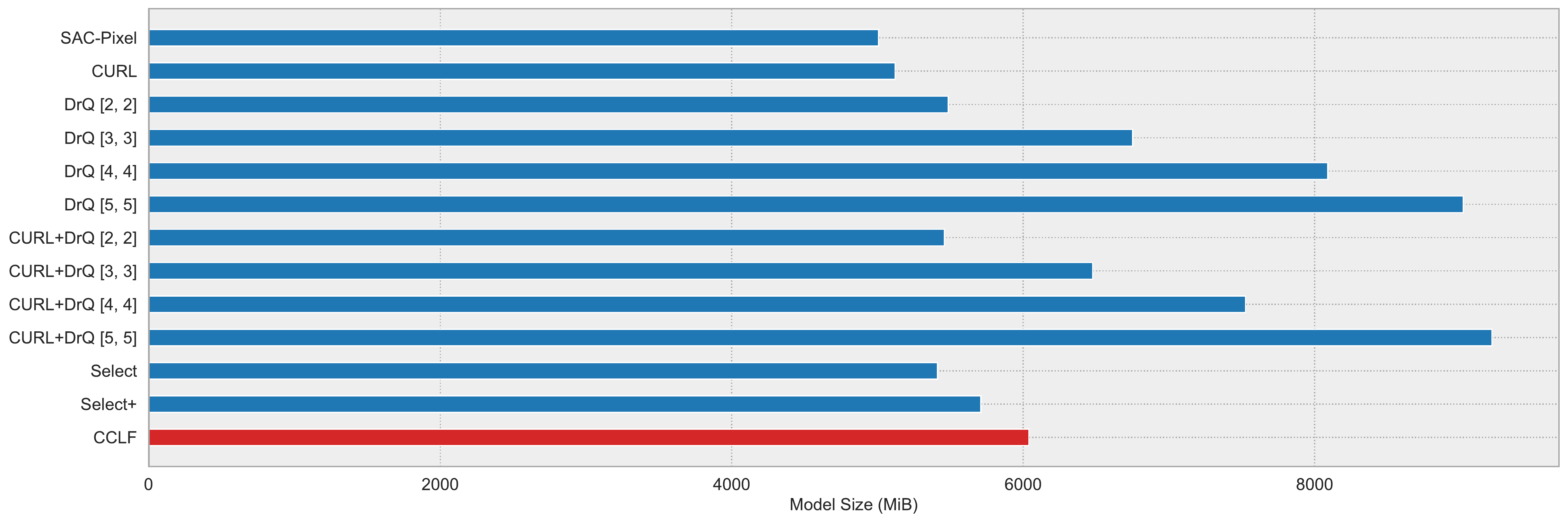} 
	\caption{Comparison of model sizes during training}
	\label{model_size}
\end{figure*}

\paragraph{Model Sizes.}
Figure \ref{model_size} shows the average model size across different models during training. As we employ five servers with different configurations, we take the mean size of the models running on different servers. It can be observed from Figure~\ref{model_size} that when increasing the number of augmentation inputs $[K, M]$ for both DrQ and CURL+DrQ, their model sizes increase substantially, which cause the increase of computational complexity as well. However, CCLF only requires relatively small model sizes and can efficiently reduce the prohibitive cost. In particular, CCLF only selects the {\bf{two}} most informative augmented inputs from the {\bf{five}} augmented inputs $[K, M]=[5,5]$ respectively for both current observations and next observations, where the amount of inputs is reduced by 60\%. On the one hand, our CCLF reduces 33\% and 34.5\% of the model sizes, respectively, compared to DrQ [5, 5] and CURL+DrQ [5, 5]. On the other hand, CCLF only sightly increases the model size by 10.1\% and 10.6\% compared to DrQ [2, 2] and CURL+DrQ [2, 2], and does not exceed DrQ [3, 3] nor CURL+DrQ [3, 3]. Therefore, we can conclude that the proposed CCLF is capable of learning efficiently by avoiding increasing the model complexity during agent learning.

\begin{figure}[h]
	\centering
	\includegraphics[width=0.45\textwidth]{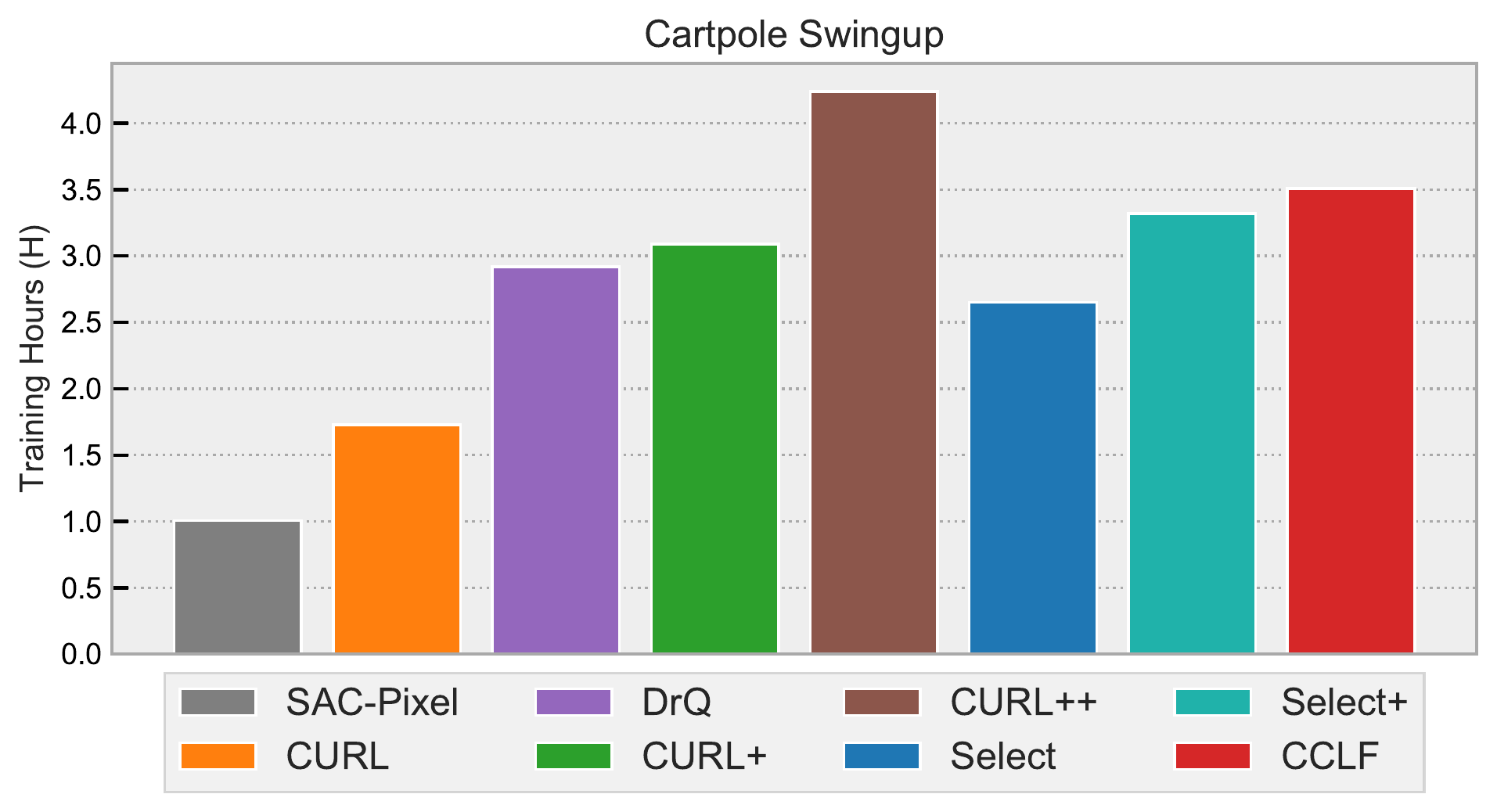} 
	\caption{Average training hours on the Cartpole-Swingup task by different models.}
	\label{fig:training_hours}
\end{figure}

\paragraph{Training Time.} To compare the computational complexity by model training time, we average on the Cartpole-Swingup task across 6 runs. Figure~\ref{fig:training_hours} shows the average training time required by different models. It can be observed that increasing the number of augmented inputs per sample will result in a significant increase (around 33\%) of training time by comparing CURL+ (2 inputs) and CURL++ (5 inputs). However, CCLF can reduce about 50\% of the increased time caused by CURL++. Overall, it only costs a moderate level of training time increase compared to both DrQ and CURL+. Therefore, we may conclude that our proposed CCLF does not require too much additional running time and can efficiently reduce the unnecessary computational cost by avoiding injecting too many inputs.


\subsection{Results on Atari Games}
\label{app:rainbow_results}

\begin{table*}[t]
	\small
	\begin{tabular}{@{}p{3.3cm}|p{0.06\textwidth}p{0.06\textwidth}|p{0.06\textwidth}p{0.07\textwidth}p{0.1\textwidth}p{0.08\textwidth}p{0.06\textwidth}p{0.05\textwidth}p{0.06\textwidth}@{}}
		\toprule
		Atari100K Games& Human   & Random & SimPLe          & OTRainbow      & Eff.  Rainbow  & Eff. DQN       & CURL           & DrQ            & CCLF            \\* \midrule
		ALIEN           & 7127.7  & 227.8  & 616.9           & 824.7          & 739.9          & 558.1          & 558.2          & 771.2          & \textbf{920.0}   \\
		AMIDAR          & 1719.5  & 5.8    & 88.0            & 82.8           & \textbf{188.6} & 63.7           & 142.1          & 102.8          & 154.7            \\
		ASSAULT         & 742.0   & 222.4  & 527.2           & 351.9          & 431.2          & 589.5          & 600.6          & 452.4          & \textbf{612.4}   \\
		ASTERIX         & 8503.3  & 210.0  & \textbf{1128.3} & 628.5          & 470.8          & 341.9          & 734.5          & 603.5          & 708.8            \\
		BANK HEIST      & 753.1   & 14.2   & 34.2            & \textbf{182.1} & 51.0           & 74.0           & 131.6          & 168.9          & 36.0             \\
		BATTLE ZONE    & 37187.5 & 2360.0  & 5184.4           & 4060.6  & 10124.6 & 4760.8 & \textbf{14870.0} & 12954.0 & 5775.0  \\
		BOXING          & 12.1    & 0.1    & \textbf{9.1}    & 2.5            & 0.2            & -1.8           & 1.2            & 6.0            & 7.4              \\
		BREAKOUT        & 30.5    & 1.7    & \textbf{16.4}   & 9.8            & 1.9            & 7.3            & 4.9            & 16.1           & 2.7              \\
		CHOPPER COMMAND & 7387.8  & 811.0  & \textbf{1246.9} & 1033.3         & 861.8          & 624.4          & 1058.5         & 780.3          & 765.0            \\
		CRAZY CLIMBER  & 35829.4 & 10780.5 & \textbf{62583.6} & 21327.8 & 16185.3 & 5430.6 & 12146.5          & 20516.5 & 7845.0  \\
		DEMON ATTACK    & 1971.0  & 152.1  & 208.1           & 711.8          & 508.0          & 403.5          & 817.6          & 1113.4         & \textbf{1360.9}  \\
		FREEWAY         & 29.6    & 0.0    & 20.3            & 25.0           & \textbf{27.9}  & 3.7            & 26.7           & 9.8            & 22.6             \\
		FROSTBITE       & 4334.7  & 65.2   & 254.7           & 231.6          & 866.8          & 202.9          & 1181.3         & 331.1          & \textbf{1401.0}  \\
		GOPHER          & 2412.5  & 257.6  & 771.0           & 778.0          & 349.5          & 320.8          & 669.3          & 636.3          & \textbf{814.7}   \\
		HERO            & 30826.4 & 1027.0 & 2656.6          & 6458.8         & 6857.0         & 2200.1         & 6279.3         & 3736.3         & \textbf{6944.5}  \\
		JAMESBOND       & 302.8   & 29.0   & 125.3           & 112.3          & 301.6          & 133.2          & \textbf{471.0} & 236.0          & 308.8            \\
		KANGAROO        & 3035.0  & 52.0   & 323.1           & 605.4          & 779.3          & 448.6          & 872.5          & \textbf{940.6} & 650.0            \\
		KRULL           & 2665.5  & 1598.0 & \textbf{4539.9} & 3277.9         & 2851.5         & 2999.0         & 4228.6         & 4018.1         & 3975.0           \\
		KUNG FU MASTER & 22736.3 & 258.5   & \textbf{17257.2} & 5722.2  & 14346.1 & 2020.9 & 14307.8          & 9111.0  & 12605.0 \\
		MS PACMAN       & 6951.6  & 307.3  & \textbf{1480.0} & 941.9          & 1204.1         & 872.0          & 1465.5         & 960.5          & 1397.5           \\
		PONG            & 14.6    & -20.7  & \textbf{12.8}   & 1.3            & -19.3          & -19.4          & -16.5          & -8.5           & -17.3            \\
		PRIVATE EYE     & 69571.3 & 24.9   & 58.3            & 100.0          & 97.8           & \textbf{351.3} & 218.4          & -13.6          & 100.0            \\
		QBERT           & 13455.0 & 163.9  & \textbf{1288.8} & 509.3          & 1152.9         & 627.5          & 1042.4         & 854.4          & 953.8            \\
		ROAD RUNNER     & 7845.0  & 11.5   & 5640.6          & 2696.7         & 9600.0         & 1491.9         & 5661.0         & 8895.1         & \textbf{11730.0} \\
		SEAQUEST        & 42054.7 & 68.4   & \textbf{683.3}  & 286.9          & 354.1          & 240.1          & 384.5          & 301.2          & 550.5            \\
		UP N DOWN       & 11693.2 & 533.4  & 3350.3          & 2847.6         & 2877.4         & 2901.7         & 2955.2         & 3180.8         & \textbf{3376.3}          \\* \bottomrule
	\end{tabular}
	\caption{Average episode returns on each of 26 Atari games at 100K training steps, across 4 random runs. In each game, the highest score is bold, where the scores of baseline models are listed in both DrQ and CURL papers. The proposed CCLF demonstrates better overall performance on 8 out of 26 games.}
	\label{rainbow_results}
\end{table*}

We present the results on Arari Games at 100K training steps in Table \ref{rainbow_results}. Overall, the proposed CCLF achieves the state-of-the-art performance on 8 out of 26 games. In particular, we include the following baselines to compare against:
\begin{itemize}
	\item Random Agent and Human Performance (Human).
	\item SimPLe \cite{kaiser2019model}, a model-based algorithm.
	\item OverTrained Rainbow (OTRainbow) \cite{kielak2019recent}.
	\item Data-Efficient Rainbow (Eff. Rainbow) \cite{van2019use}.
	\item Efficient DQN \cite{mnih2013playing}.
	\item CURL \cite{laskin2020curl}.
	\item DrQ \cite{yarats2020image}.
\end{itemize}
For consistency, we directly use the average scores of all baseline models that are recorded in DrQ and CURL papers. It can be seen from Table \ref{rainbow_results} that CCLF can improve over CURL and DrQ on 11 and 18 out of 26 games, with about 18\% and 39\% performance improvements on average respectively. In particular, we have obtained 2.07$\times$, 1.66$\times$, and 1.65$\times$ mean scores than CURL on the Road Runner, Demon Attack, and Alien games. Compared to DrQ, CCLF outperforms it with 2.30$\times$, 1.86$\times$, and 1.83$\times$ mean scores on the Freeway, Hero, and Seaquest games. These results have demonstrated the desired sample-efficient capability of the proposed CCLF.
In addition, CCLF even achieves superhuman performance on Krull and Road Runner tasks. Although CCLF attains state-of-the-art in 8 games, there are still some gaps, compared to the human performances and the top performing model-based method SimPLe on the other tasks. This shortcoming can also be found in both DrQ and CURL results, where further improvement towards human-level performance is desired.

\subsection{Additional Discussion on MiniGrid}
\label{app:A2C_results}
As shown in Figure~\ref{fig:A2Cresults} from the main paper, CCLF can significantly improve the sample efficiency as well as the ultimate learning performances, by extending both A2C and RE3 algorithms. In particular, incorporating CCLF directly on A2C is sufficient to outperform RE3 by reaching optimal performance levels faster with even higher scores in all three tasks. Meanwhile, integrating CCLF to RE3 can further improve the sample efficiency. In the DoorKey-8x8 task, our agent started to obtain non-trivial scores at around 700K steps, while conventional A2C has failed even at 2400K step and RE3 started to improve only at around 1200K steps.
As RE3 has been empirically proven more effective than other curiosity-driven methods (ICM \cite{pathak2017curiosity} and RND\cite{burda2018exploration}), we can conclude that our CCLF is sample-efficient and is capable of encouraging exploration effectively in the MiniGrid environments.

\subsection{Effectiveness of the Proposed Components}
\label{app:ablation}
One might wonder if the proposed CCLF benefits mainly from one or several curiosity-based components in practice. Hence, we empirically examine the effectiveness of all possible (15) combinations of the proposed four components on the Cartpole task from the DMC suite, averaging by 6 random runs. 
In this task, agents need to swing up a pole over the cart by continuously moving the cart around. 
\begin{figure*}[h!]
	\centering
	\begin{subfigure}{0.33\textwidth}
		\includegraphics[width=1\textwidth]{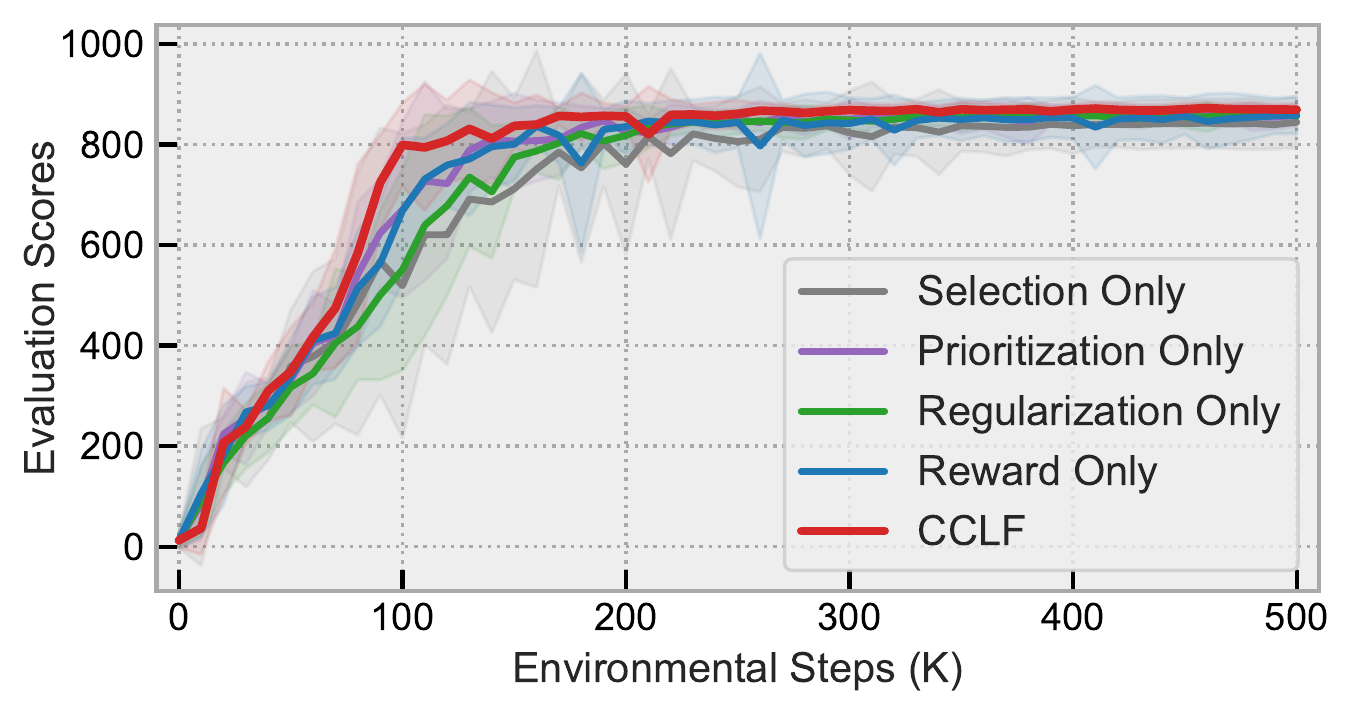} 
		\caption{Add One Component}
	\end{subfigure}
	\begin{subfigure}{0.33\textwidth}
		\includegraphics[width=1\textwidth]{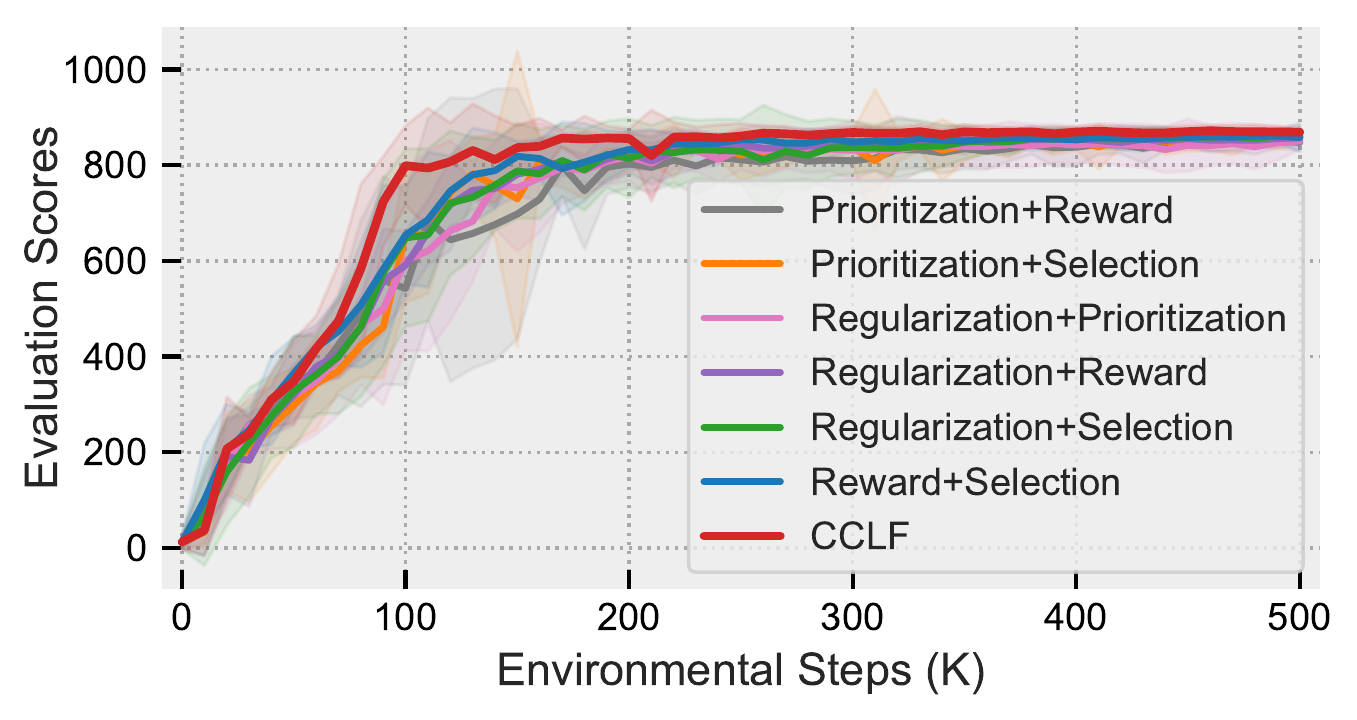} 
		\caption{Add Two Components}
	\end{subfigure}
	\begin{subfigure}{0.33\textwidth}
		\includegraphics[width=1\textwidth]{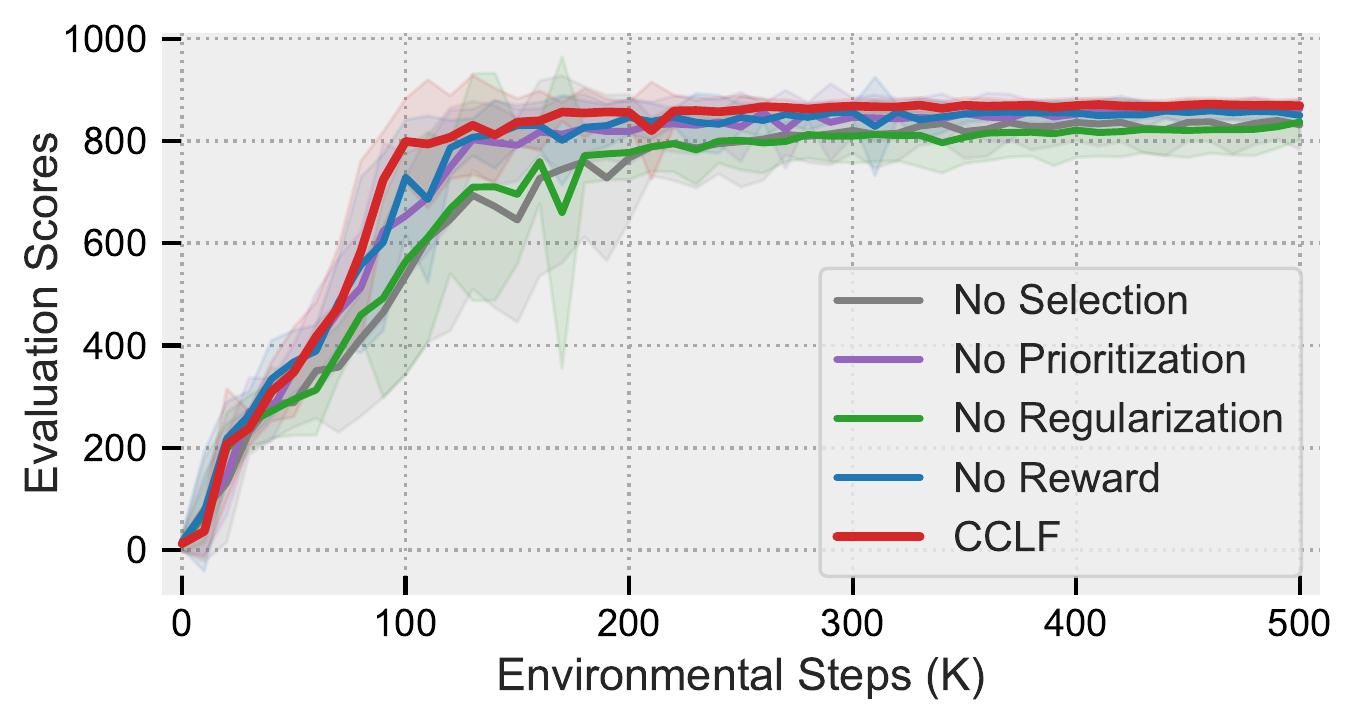} 
		\caption{Add Three Components (Remove One)}
	\end{subfigure}
	\caption{An investigation of the effectiveness of all curiosity-based components on the Cartpole-Swingup task. Adding one component alone or any two components is not sufficient for learning improvement, but instead can cause extra instability and complexity. Removing reward or prioritization has less impact than removing regularization or selection, but it is still less competitive than using all four components. }
	\label{fig:ablation}
\end{figure*}

\subsubsection{Adding One Component}
We respectively incorporate each curiosity-based component (Selection Only, Prioritization Only, Regularization Only, and Reward Only) into the SAC base algorithm and compare their performances again our full CCLF. It can be observed from Figure~\ref{fig:ablation}(a) that all models approximately obtain the same performances at 500K regimes. For the sample efficiency, all four curves are significantly below the proposed CCLF at 100K. Meanwhile, there exists some sudden increase or decrease in the learning curves of all four models, indicating the occurrence of instability. Therefore, each component alone cannot achieve the desired sample efficiency and can even cause unstable learning qualitatively. In contrast, our collaborative CCLF can navigate all components together, which demonstrates effective collaboration.

\begin{table}[h!]
	\begin{tabular}{@{}lll@{}}
		\toprule
		Model           & 100K Step Score     & 500K Step Score    \\ \midrule
		Regularization Only & 551$\pm$146         & 857$\pm$10         \\
		Selection Only     & 561$\pm$181         & 837$\pm$38         \\
		Reward Only     & 668$\pm$105         & 858$\pm$27         \\
		Prioritization Only & 670$\pm$127         & 858$\pm$16         \\
		CCLF            & \textbf{799$\pm$61} & \textbf{869$\pm$9} \\ \bottomrule
	\end{tabular}
	\caption{Results for the sample efficiency at 100K environment steps and asymptotic performance at 500K environment steps by adding only one proposed component. }
	\label{ablation_table1}
\end{table}

In addition, we further compare and analyze the sample efficiency and ultimate performance quantitatively according to the results in Table \ref{ablation_table1}. For the sample efficiency at 100K environment steps, Reward-Only and Prioritize-Only obtain similar performances, which outperform Regularize-Only and Select-Only. However, our proposed CCLF achieves even a 1.19 $\times$ mean score at 100K steps compared to the best performance among these four models. Meanwhile, we can observe that the standard deviations of these four models are much higher than our proposed CCLF at both 100K and 500K environment steps. This further implies that each component alone cannot resolve the instability issue that occurred in the learning process. In particular, Select-Only has the highest standard deviation because it selects the most challenging inputs that agents cannot accurately predict for learning. However, as the learning process is not properly adapted to capture the novel knowledge contained in the selected inputs, the instability issues cannot be avoided. CCLF collaboratively adapts the learning process with all four RL components and therefore its standard deviation is the lowest, where the agents can learn with contrastive curiosity. From the perspective of the ultimate performance, the proposed CCLF is also the best compared to the other four models, indicating an improved learning capability as the result of the efficient collaboration of all RL components as well.

\subsubsection{Adding Two Components}
Two proposed components are incorporated into the base algorithm in this sub-section. It can be observed from Figure~\ref{fig:ablation}(b) that all models approximately obtain the same performances at 500K regimes. For the sample efficiency, all six curves are significantly below the proposed CCLF at 100K. Among the six experimented models, Reward+Selection performs the best while Prioritization+Reward learns the worst. It seems that regularization and selection are more important to attain the desired performance. However, only incorporating these two components (Regularization+Selection) may instead introduce some instability, which should be addressed by the remaining two components as a whole. In contrast, our collaborative CCLF can navigate all components together, which demonstrates effective collaboration.

\begin{table}[h!]
	\begin{tabular}{@{}lll@{}}
		\toprule
		Model           & 100K Score     & 500K Score    \\ \midrule
		Prioritization+Reward & 542$\pm$147         & 848$\pm$12         \\
		Prioritization+Selection & 655$\pm$104         & 859$\pm$9         \\
		Regularization+Prioritization & 601$\pm$137         & 851$\pm$13         \\
		Regularization+Reward & 591$\pm$52         & 857$\pm$12        \\
		Regularization+Selection & 648$\pm$135         & 861$\pm$11         \\
		Reward+Selection & 654$\pm$74         & 858$\pm$22         \\
		CCLF            & \textbf{799$\pm$61} & \textbf{869$\pm$9} \\ \bottomrule
	\end{tabular}
	\caption{Results for the sample efficiency at 100K environment steps and asymptotic performance at 500K environment steps by adding two proposed components. }
	\label{ablation_table2}
\end{table}

Quantitatively, we compare and analyze the sample efficiency and ultimate performance based on the results in Table \ref{ablation_table2}. For the sample efficiency at 100K environment steps, Prioritization+Selection, Regularization+Selection and Reward+Selection obtain similar results, which outperform the others with Prioritization+Reward being the worst. However, our proposed CCLF achieves even a 1.22$\times$ mean score at 100K steps compared to the best performance among these six models. Meanwhile, we can observe that the standard deviations of these models except for Regularization+Reward are much higher than our proposed CCLF at 100K environment steps. It implies that CCLF can stably adapt the learning process by the contrastive curiosity that seamlessly connects all four components.

\subsubsection{Adding Three Components}
We respectively remove one component from the full CCLF and present the learning performances in Figure~\ref{fig:ablation}(c). In particular, removing only prioritization (No Prioritization) or reward (No Reward) only results in a slight downward shift in the learning curves while No Selection and No Regularization show significant performance downgrade in both sample efficiency and learning capabilities. It indicates that all four components are necessarily important to attain state-of-the-art results, where regularization and selection are more important than prioritization and reward. 

\begin{table}[h!]
	\begin{tabular}{@{}lll@{}}
		\toprule
		Model           & 100K Score     & 500K Score    \\ \midrule
		No Selection & 536$\pm$141         & 832$\pm$35         \\
		No Prioritization     & 653$\pm$89        & 863$\pm$10         \\
		No Regularization     & 565$\pm$161         & 837$\pm$25         \\
		No Reward & 729$\pm$82         & 859$\pm$17         \\
		CCLF            & \textbf{799$\pm$61} & \textbf{869$\pm$9} \\ \bottomrule
	\end{tabular}
	\caption{Results for the sample efficiency at 100K environment steps and asymptotic performance at 500K environment steps by adding three (removing one) proposed components. }
	\label{ablation_table3}
\end{table}

Quantitatively, we also compare and analyze mean scores at 100K and 500K steps, as shown in Table \ref{ablation_table3}. For the sample efficiency at 100K environment steps, No Selection and No Regularization result in 33\% and 29\% return decreases compared to the full model; meanwhile, their standard deviations are much higher, indicating the occurrence of instability. Similarly, these two models have caused performance downgrades at even 500K steps, with larger standard deviations as well.
As a result, we believe regularization and selection play more important roles in CCLF, but it still requires the four components to work collaboratively to obtain the desired sample efficiency and learning performances.

\end{appendices}
	
\end{document}